\documentclass[journal]{IEEEtran}

\IEEEoverridecommandlockouts                              

\usepackage{graphicx} 
\usepackage{caption}
\usepackage{subcaption}
\graphicspath{{./images}}

\usepackage{enumitem}
\usepackage{color}
\usepackage{xcolor,colortbl}
\usepackage{multicol,multirow}
\definecolor{Gray}{gray}{0.85}
\definecolor{LightCyan}{rgb}{0.88,1,1}
\newlength\savewidth\newcommand\shline{\noalign{\global\savewidth\arrayrulewidth
		\global\arrayrulewidth 1pt}\hline\noalign{\global\arrayrulewidth\savewidth}}

\usepackage{pifont}         
\newcommand{\cmark}{\ding{51}}
\newcommand{\xmark}{\ding{55}}

\title{
Robust Roadside Perception:
an Automated Data Synthesis Pipeline Minimizing Human Annotation
}

\author{Rusheng Zhang$^{1}$, Depu Meng$^{1}$, Lance Bassett$^{2}$, Shengyin Shen$^{3}$, Zhengxia Zou$^{4}$ and Henry X. Liu$^{1,5*}$
\thanks{This work is part of US DoT Smart Intersection Project.}
\thanks{$^{1}$Rusheng Zhang, Depu Meng, and Henry X. Liu are with the Department of Civil and Environmental Engineering, University of Michigan, Ann Arbor.}
\thanks{$^{2}$Lance Bassett is with the Department of Electrical Engineering and Computer Science, University of Michigan, Ann Arbor.}
\thanks{$^{3}$Shengyin Shen is with University of Michigan Transportation Research Institute.}
\thanks{$^{4}$Zhengxia Zou is with the Beihang University.}
\thanks{$^{5}$ Henry X. Liu is also with Mcity.}
\thanks{$^{*}$Corresponding author - Henry X. Liu (henryliu@umich.edu).}
}

\begin{document}

\maketitle

\begin{abstract}
Recently, advancements in vehicle-to-infrastructure communication technologies have elevated the significance of infrastructure-based roadside perception systems for cooperative driving. This paper delves into one of its most pivotal challenges: data insufficiency. The lacking of high-quality labeled roadside sensor data with high diversity leads to low robustness, and low transfer-ability of current roadside perception systems. In this paper, a novel solution is proposed to address this problem that creates synthesized training data using Augmented Reality. A Generative Adversarial Network is then applied to enhance the reality further, that produces a photo-realistic synthesized dataset that is capable of training or fine-tuning a roadside perception detector which is robust to different weather and lighting conditions.

Our approach was rigorously tested at two key intersections in Michigan, USA: the Mcity intersection and the State St./Ellsworth Rd roundabout. The Mcity intersection is located within the Mcity test field, a controlled testing environment. In contrast, the State St./Ellsworth Rd intersection is a bustling roundabout notorious for its high traffic flow and a significant number of accidents annually.
Experimental results demonstrate that detectors trained solely on synthesized data exhibit commendable performance across all conditions. Furthermore, when integrated with labeled data, the synthesized data can notably bolster the performance of pre-existing detectors, especially in adverse conditions.
\end{abstract}

\section{INTRODUCTION}
Recently, with the rapid development in vehicle-to-infrastructure (V2I) communications technologies, the infrastructure-based perception system to support autonomous driving has attracted significant attention. Sensors installed on the roadside detect vehicles in the region-of-interest in real-time, and forward the perception results to Connected Automated Vehicles (CAVs) with short latency via V2I communications, broadcasting Basic Safety messages defined in SAE J2735 \cite{draft2006j2735, kenney2011dedicated} or Sensor Data Sharing Message defined in SAE J3224 \cite{sae2019v2x}. The roadside sensors are usually positioned at fixed locations on the roadside, typically at elevated heights, providing a broader perspective, fewer obstructed objects, blind spots, and less environmental variability compared to the sensors onboard the vehicles. Consequently, the perception results obtained from the roadside sensors can complement the onboard perception of CAVs, resulting in a more comprehensive, consistent, and accurate understanding of the surrounding scene, particularly in complex scenarios and challenging weather and lighting conditions.  

\begin{figure}[t]
    \centering
     \begin{subfigure}[b]{0.9\linewidth}
         \centering
         \includegraphics[width=\textwidth]{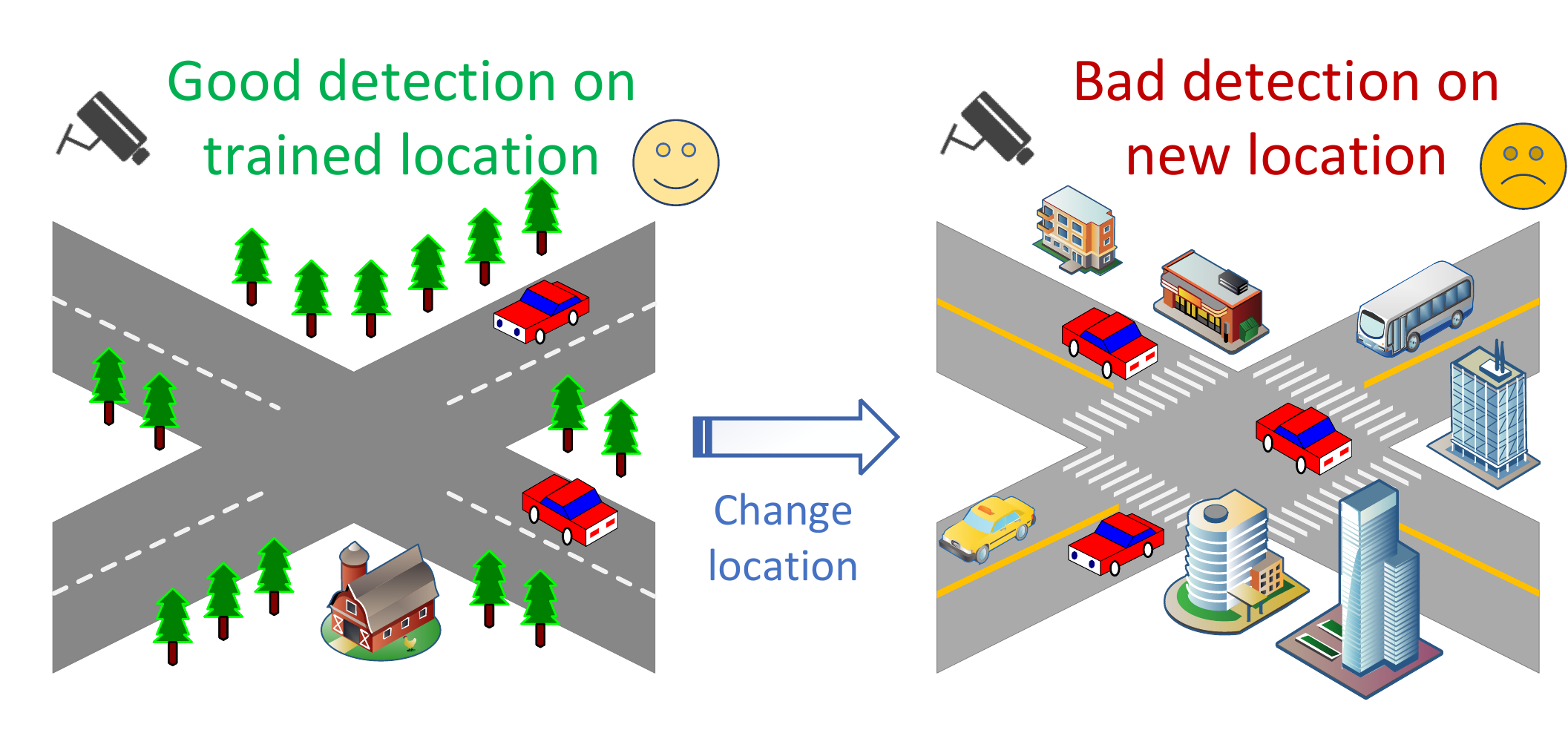}
         \caption{A detector trained in one location may perform poorly when deploying the same detector to another location.}
         \label{fig:introduction1}
     \end{subfigure}
     \\
     \hspace{1mm}
     \begin{subfigure}[b]{0.9\linewidth}
         \centering
         \includegraphics[width=\textwidth]{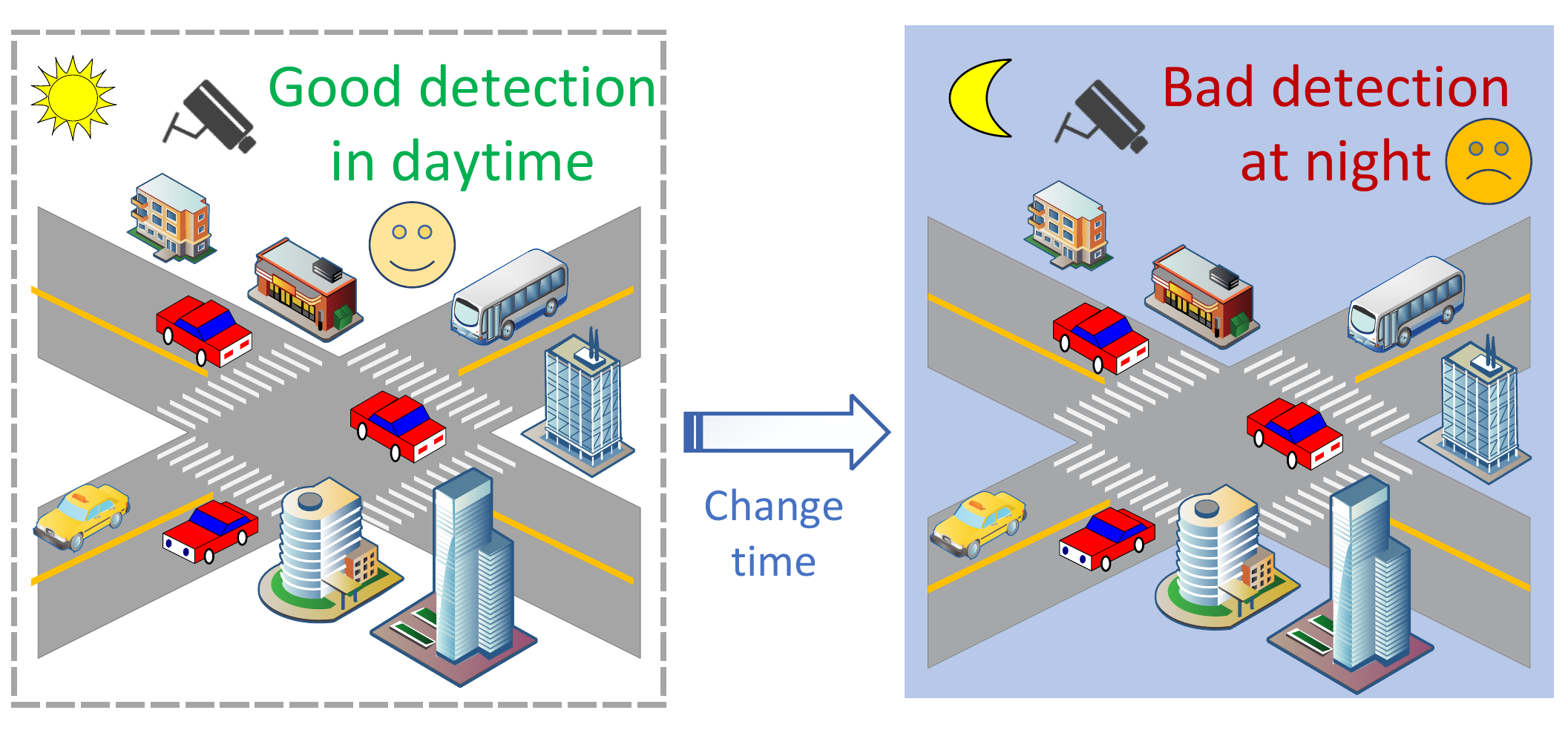}
         \caption{A detector trained with day time data may perform poorly on the same location at night.}
         \label{fig:introduction2}
     \end{subfigure}
        \caption{An illustrative figure that shows the issues current roadside perception systems face due to the data insufficiency.}
\vspace{-3mm}
        \label{fig:problem-introduction}
\end{figure}

Although it is commonly believed that roadside perception is less complex than onboard perception due to the significantly lower environmental variability and fewer occluded objects, roadside perception presents its own unique set of challenges. One of the most crucial challenges is data insufficiency, specifically the lack of high-quality and diverse labeled sensor data collected from the roadside. Acquiring roadside data with a sufficient level of diversity (through the deployment of numerous sensors along the roadside) is expensive compared to onboard perception, primarily due to the high cost of installation. Furthermore, obtaining large quantities of labeled data is even more costly, given the high expense associated with labeling. Currently, high-quality labeled roadside perception data are generally obtained from few locations with limited environmental diversity.  
 
The aforementioned challenge of data insufficiency presents significant practical issues in real-world deployment. Figure \ref{fig:problem-introduction} provides illustrative examples on these issues. In Figure \ref{fig:introduction1}, the performance of the detector trained on data from one location is severely compromised when applied to a new location. Similarly, in Figure \ref{fig:introduction2}, the training dataset lacks images taken at night, resulting in poor performance during nighttime, even at the same location. Clearly, these challenges impede the widespread deployment of roadside perception systems. Moreover, since roadside perception is considered a supplementary and enhancing approach to onboard vehicle detection, it is expected to have higher requirements for robustness and accuracy compared to onboard perception. Consequently, such demanding nature of roadside perception exacerbates the aforementioned challenge of data-insufficiency.
 
In this paper, we propose an automated pipeline that generates photo-realistic synthesized dataset for roadside perception using Augmented Reality (AR) \cite{billinghurst2015survey} and Generative Adversarial Network (GAN) \cite{creswell2018generative}. 
This synthesized dataset can be used either to train a high-performance roadside perception system, or to fine-tune an existing detector with minimum human labeling effort. 
Since the data synthesizing quality could be further improved with more unlabeled data, this work can be considered a step forward to large-scale real-world deployment. 
 
 The contributions of this paper are as follows:
 \begin{enumerate}
    \item Propose an AR rendering pipeline for a roadside perception system, including camera pose estimation, realistic vehicle positioning and heading simulation, and AR rendering. This pipeline generates physically realistic images with annotations. 
     \item Propose a GAN based reality enhancement strategy that processes the physical realistic images obtained from AR and converts them to photo-realistic images.
     \item Report a thorough field evaluation of the model obtained from the aforementioned pipeline under different lighting and weather conditions that demonstrates the viability of this method in real-world, large-scale deployment.
 \end{enumerate}

\section{Related Works}
\subsection{Roadside Camera for Vehicle Detection}
The existence of roadside sensor-based surveillance systems can be traced back as early as $1986$. Early systems aim at traffic monitoring and abnormal behavior detection \cite{datondji2016survey}. These systems consist of single or multiple cameras mounted at a high elevated position. A large number of algorithms have been explored during the last decade for roadside vehicle detection with cameras, to name a few, background subtraction \cite{furuya2014road}, frame difference \cite{messelodi2005computer}, feature-based detection \cite{saunier2006feature}, KanadeLucas-Tomasi tracking \cite{li2016robust},  cascading classifiers \cite{faisal2020automated} and many more \cite{datondji2016survey}. Recently, deep learning based detection has become the trend \cite{aboah2021vision, zhang2022design, zou2022real,zhang2023msight}.

\subsection{Training on Synthesized Data}
Acquiring a substantial quantity of data is a costly endeavor, and obtaining annotations further escalates the expenses. As a result, the insufficiency of data presents a widespread challenge in the field of deep learning. To address this challenge, several synthesized datasets and simulation tools,
such as Carla \cite{dosovitskiy2017carla}, AirSim \cite{shah2018airsim}, SYNTHIA \cite{7780721}, GTA5 \cite{richter2016playing} and VIPER \cite{richter2017playing},
are developed, so that researchers can have access to sufficient annotated data. However, the neural networks trained on these synthesized datasets often perform poorly in the real-world. 

There have been many proposals on using synthesized data together with real-world data to mitigate such effects \cite{seib2020mixing}. For example, \cite{tremblay2018training} trains neural networks with synthesized data generated with domain randomization and shows that the network yields better results using the augmented data than with real-world data alone. Generative Adversarial Networks (GANs) \cite{creswell2018generative} represent another promising approach that has recently been explored for data augmentation due to their potential to generate photo-realistic images. In a study conducted by \cite{perez2017effectiveness}, various data augmentation schemes were evaluated, and the use of GANs to generate images in different styles was proposed. \cite{yang2020surfelgan} introduced SurfelGAN, a method that synthesizes realistic images reconstructed from actual sensor data. Additionally, GANs have been utilized by \cite{9565008} to translate clear images into different weather conditions, thereby enhancing the robustness of autonomous vehicles in adverse weather. Finally, \cite{zhan2020adversarial} proposed an image composition method that employs both local and global discriminators to achieve realistic shadow and texture effect.

In the field of autonomous driving, data augmentation with image composition has recently become an interesting topic. Several researches focus on data augmentation for onboard vehicle perception. \cite{chen2021geosim} proposes GeoSim, a geometry-aware image composition process that can augment existing image with dynamic objects. \cite{wangcadsim} proposes CADSim that recovers photo-realistic 3D traffic participant model from sparse sensor data, to create a large set of 3D model library which can be further used in data augmentation and simulations. These method can generate photo-realistic augmented data for perception and other safety-critic downstream applications. 
While these methods can potentially be applied directly, their need for 3D awareness makes them best suited for scenarios where both point cloud and camera images are accessible, especially given the current limited availability of infrastructure data that combines both.

Given the static nature of the background in roadside sensor data, the process of image composition becomes considerably simpler. Unlike onborad dynamic scenes that are replete with challenges such as occlusion, variable lighting, and unpredictable motion, static backgrounds afford a stable canvas on which augmented elements can be added with relative ease. This simplicity has led to the emergence of various methods capable of accomplishing such tasks. In fact, many of the current methods have advanced to a degree that their capabilities surpass the needs of simple roadside data augmentation. It's important to note that the main intent of this paper isn’t to claim superiority in this specific domain of data augmentation. Instead, our focus is to establish a methodology pipeline that demonstrates the efficacy of data synthesis in the context of roadside perception. By leveraging the strengths of existing technologies and adapting them for this niche, we aim to open new avenues in the field of roadside cooperative perception.

\section{Methodology}
\subsection{Hardware Setup}
\vspace{-3mm}
\begin{figure}[ht]
    \centering
    \begin{subfigure}[b]{.47\linewidth}
         \centering
         \includegraphics[height=.84\textwidth]{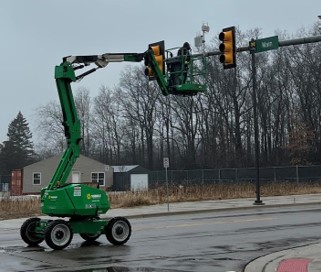}
         \caption{One of the mast arm in Mcity where the camera is being installed.}
         \label{fig:installation1}
     \end{subfigure}
     \hfill
     \begin{subfigure}[b]{.47\linewidth}
         \centering
         \includegraphics[height=.75\textwidth]{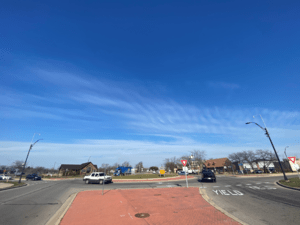}
         \caption{The roundabout at the intersection of Ellsworth St. and State St. in Ann Arbor, Michigan, USA}
         \label{fig:installation2}
     \end{subfigure}
    \caption{Two sets of cameras are leveraged at two different locations in this paper. Four cameras are installed at an intersection in Mcity, facing four approaches respectively (Figure \ref{fig:installation1}) and another four cameras are installed at four corners of a two-lane roundabout (Figure \ref{fig:installation2}).}
\vspace{-3mm}
    \label{fig:installation}
\end{figure}
The experiments reported in this paper are conducted in two location, one intersection in Mcity, and another roundabout at the intersection of Ellsworth St. and State St. in Ann Arbor, Michigan, USA. Mcity is the world's first controlled environment specifically designed to test the performance and safety of connected and automated vehicle technologies \cite{mcity}. Four pinhole cameras are installed on the mast arms of the intersection at Mcity, each facing one approach. The two-lane roundabout is located at Ellsworth St. and State St. in Ann Arbor, Michigan, USA. Four pinhole cameras are installed at the four corner of the roundabout, recording real car flow. Figure \ref{fig:installation} shows images of these two experimental position.

\subsection{Core Idea}
\begin{figure*}[ht]
    \centering
    \includegraphics[width=.95\textwidth]{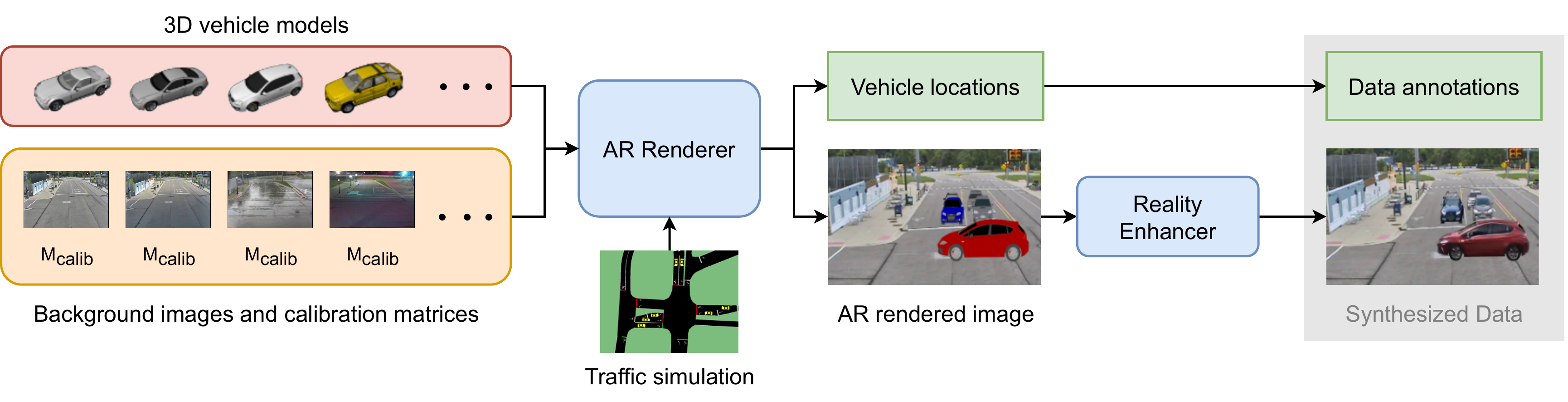}
    \vspace{-2mm}
    \caption{Data synthesizing pipeline to generate realistic data: an AR renderer renders 3D model onto the real background with traffic simulation data, a GAN-based reality enhancer is then applied to make the rendered vehicle photo-realistic.}
    \label{fig:data-synthesize-pipeline}
    \vspace{-3mm}
\end{figure*}

Our solution to create a roadside perception system includes two main steps: 
\begin{enumerate}
    \item a data synthesizing pipeline that generates labeled training data
    \item deep learning based vehicle detection and localization model trained with the synthesized data
\end{enumerate}
 The data synthesis pipeline renders realistic vehicles onto the background images obtained from the cameras and simultaneously generates the corresponding annotations.  The components in this step are discussed in section \ref{subsection:data-synthesize}, \ref{subsection:pose-estimation}, \ref{subsection:augmented-reality}, and \ref{subsection:reality-enhancement}. 
 
 The synthesized data are used to train a YOLOX detector \cite{abs-2107-08430}.
 The vehicle detection pipeline utilizes the trained YOLOX detector and generates 3D vehicle location information. The components in this step are discussed in section \ref{subsection:detection-and-localization}.

\subsection{The Data Synthesizing Pipeline}
\label{subsection:data-synthesize}

Figure \ref{fig:data-synthesize-pipeline} illustrates the overall data synthesis pipeline responsible for generating the training dataset. Initially, vehicle locations and headings are generated with a traffic simulator. Then, an AR renderer is employed to project $3$D vehicle models onto the background images. During this rendering process, annotation information for each rendered vehicle is also generated. To bridge the disparity between our synthesized data and the real world, a reality enhancer based on GAN technology is applied to each rendered vehicle. This enhancer effectively translates the vehicles into a more realistic style, thereby reducing the gap between our synthesized data and real-world scenarios.

\begin{figure}[ht]
    \centering
    \includegraphics[width=.9\linewidth]{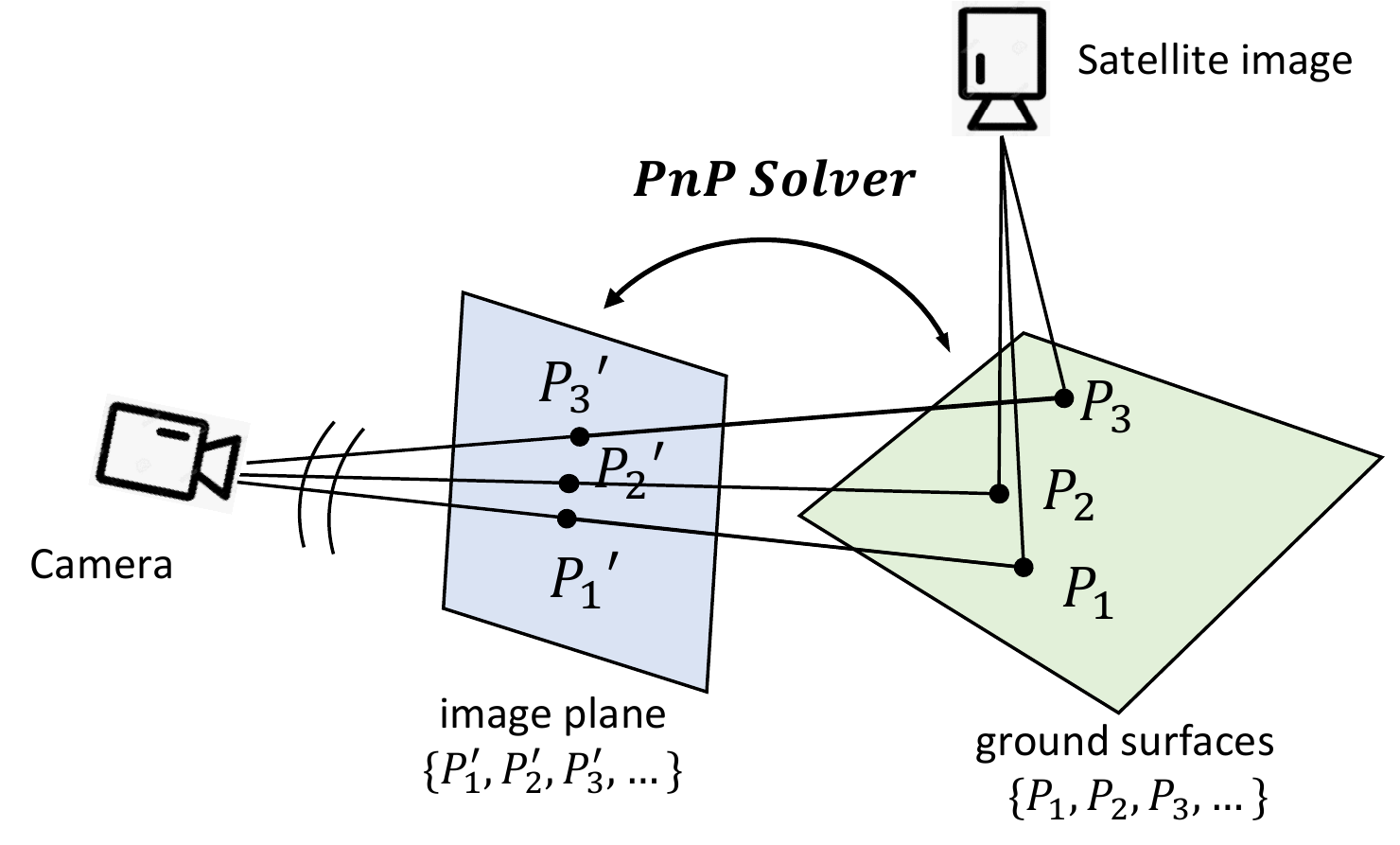}
    \caption{Illustration figure for pose estimation.}
\vspace{-3mm}
    \label{fig:pose_estimation_method}
\end{figure}

\paragraph{Background images} The background images can be easily estimated with a temporal median filter \cite{gonzalez2009digital}. One can gather the background images under different conditions to cover the variability of the background for each camera (i.e., different weather and lighting conditions).

\paragraph{Traffic simulation} A traffic simulation is performed to generate realistic vehicle trajectories; the heading and location information of the simulated vehicles is further used for AR rendering. This task is accomplished with SUMO, an open-source microscopic mobility simulator \cite{SUMO2018}. The road map information can be directly imported to SUMO from OpenStreetMap \cite{OpenStreetMap}, and constant car flows are generated for all maneuvers at the intersection. However, as SUMO only creates vehicles at the center of the lane with fixed headings, a domain randomization step is implemented to introduce variability. This involves applying a random positional and heading offset to each vehicle. The positional offset follows a normal distribution with a variance of $0.5$ meters in both vehicles' longitudinal and latitudinal directions, while the heading offset follows an uniform distribution from $-5^{\circ}$ to $5^{\circ}$.

\paragraph{3D vehicle models} The 3D vehicle models used in this work are obtained from the Shapenet repository, which is an ongoing effort to establish a richly-annotated, large-scale dataset of 3D shapes \cite{chang2015shapenet}. In total, we pick more than $200$ vehicle 3D models to yield a diverse model set. For each vehicle in SUMO simulation, a random model will be assigned and rendered onto the background images.

\subsection{Camera Pose Estimation}
\label{subsection:pose-estimation}
\begin{figure}[ht]
     \centering
     \begin{subfigure}[b]{0.47\linewidth}
         \centering
         \includegraphics[width=\textwidth]{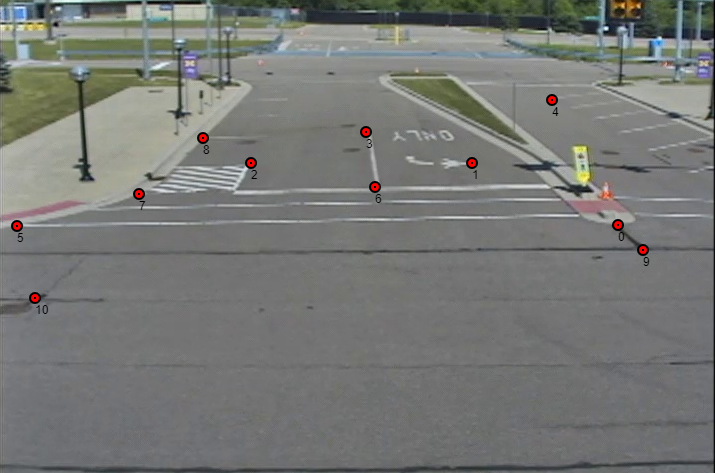}
         \caption{Landmarks selected for camera pose estimation.}
         \label{fig:landmark-photo}
     \end{subfigure}
     \hfill
     \begin{subfigure}[b]{0.45\linewidth}
         \centering
         \includegraphics[width=\textwidth]{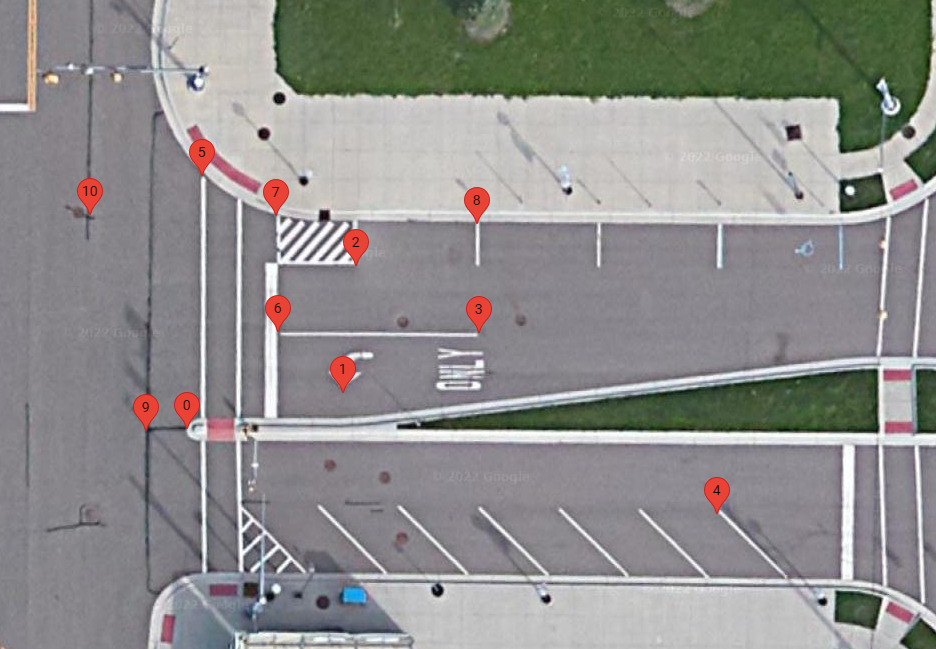}
         \caption{The corresponding landmarks from satellite image.}
         \label{fig:landmark-satelite}
     \end{subfigure}
        \caption{Pose estimation landmark selection on one of the camera (westbound approach of the intersection). }
        \label{fig:landmark-example}
    \vspace{-3mm}
\end{figure}

To correctly render 3D models onto the background images, the camera pose, including the camera rotation and camera translation in the world coordinate system, needs to be estimated. Standard camera extrinsic calibration with a large checkerboard requires on-site operation by experienced technicians, which will add complexity to the deployment pipeline, particularly in scenarios involving large-scale deployment. In this paper, we introduce a landmark based camera pose estimation method for roadside cameras that eliminates the need for field operation.  Figure \ref{fig:pose_estimation_method} provides an overview of this method. Our method considers a few landmarks, marked as $P_1, P_2, P_3, ..., P_n$ in the figure, that are both observable from the camera view and the satellite image. These landmarks provide a set of correspondences between the world coordinate system and their projections on the camera plane. 
Since camera intrinsic parameters are known, the camera pose can be solved with a Perspective-n-Point (PnP) solver using the $n$ pairs of the world-to-image correspondences obtained by these landmarks \cite{marchand2015pose}. Figure \ref{fig:landmark-example} shows the landmarks we used for one of our deployed cameras (the westbound approach view) at the intersection of Mcity. As shown in the images, we employ approximately ten landmarks for each camera to accomplish the pose estimation task. To alleviate the typically tedious nature of this process for human operators that manually match landmarks with image pixels, we have developed a user-friendly online UI interface for public use, which simplifies this task \cite{mtl_2019}.

\begin{figure*}[t]
    \centering
    \includegraphics[width=.24\textwidth, trim={1mm, 0mm, 2mm, 0mm}, clip]{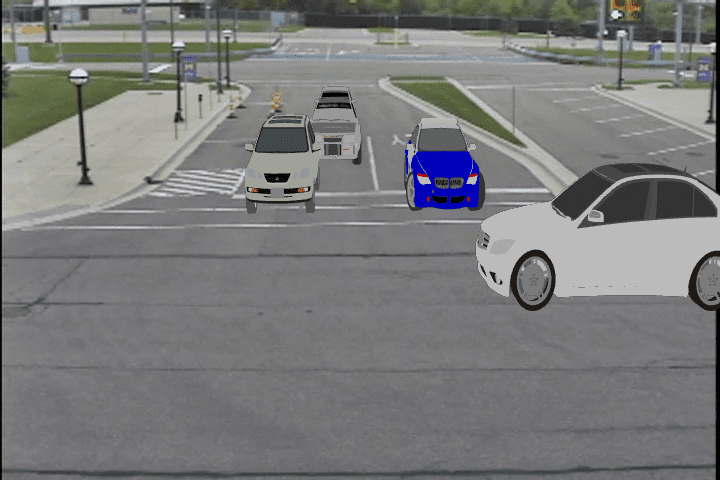}
    \includegraphics[width=.24\textwidth, trim={1mm, 0mm, 2mm, 0mm}, clip]{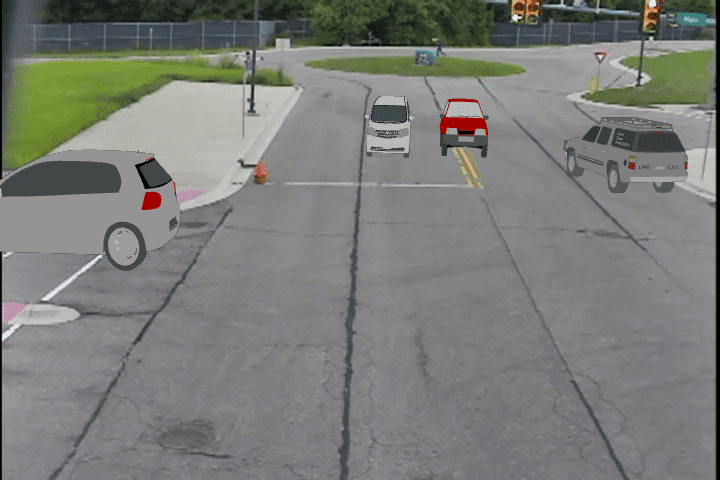}
    \includegraphics[width=.24\textwidth, trim={1mm, 0mm, 2mm, 0mm}, clip]{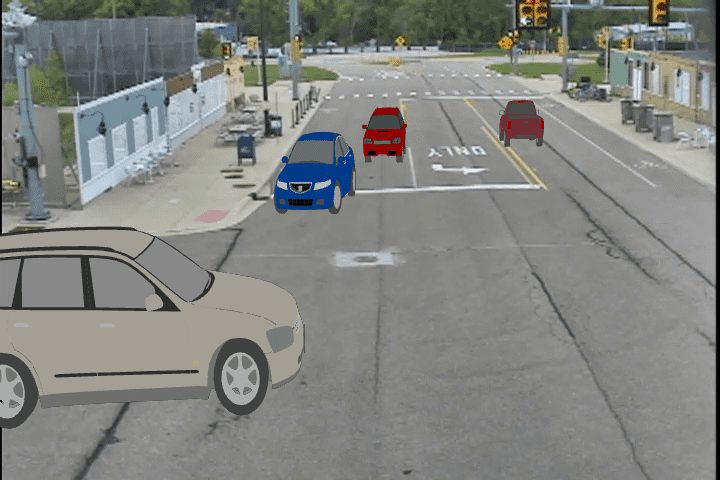}
    \includegraphics[width=.24\textwidth, trim={1mm, 0mm, 2mm, 0mm}, clip]{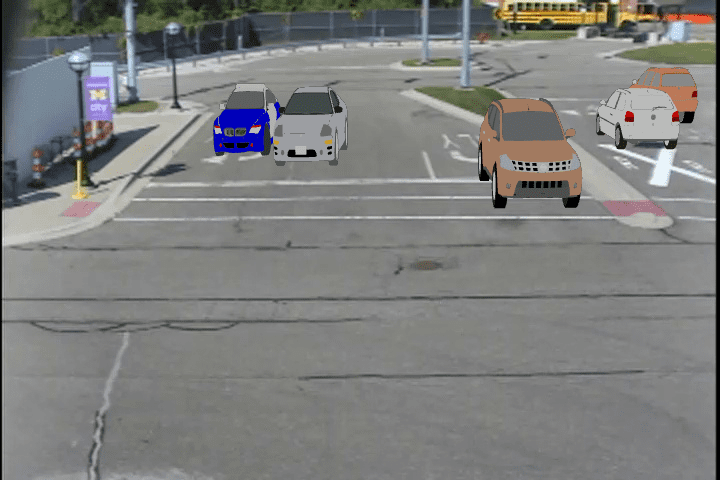} \\
    \vspace{1mm}
    \includegraphics[width=.24\textwidth, trim={1mm, 0mm, 2mm, 0mm}, clip]{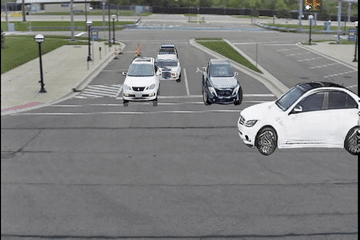}
    \includegraphics[width=.24\textwidth, trim={1mm, 0mm, 2mm, 0mm}, clip]{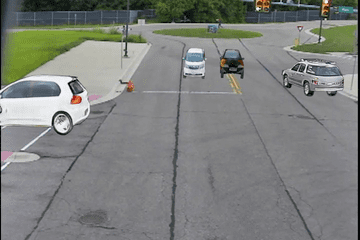}
    \includegraphics[width=.24\textwidth, trim={1mm, 0mm, 2mm, 0mm}, clip]{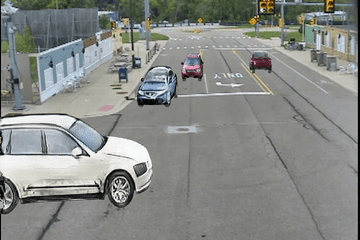}
    \includegraphics[width=.24\textwidth, trim={1mm, 0mm, 2mm, 0mm}, clip]{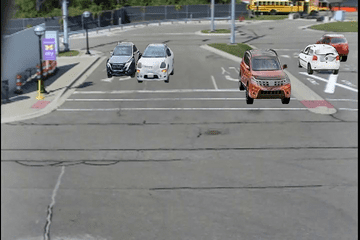} \\
    \caption{Examples of (top) Augmented Reality (AR) rendered images,
    and (bottom) the same images after reality enhancement.}
\vspace{-5mm}
    \label{fig:reality-enhancement}
\end{figure*}

\subsection{Augmented Reality Rendering}
\label{subsection:augmented-reality}
For each camera, the camera intrinsic matrix $K$ is known; the extrinsic matrix $[R|T]$ can be estimated using the method described in section \ref{subsection:pose-estimation}. Here, $R$ is a $3\times 3$ rotation matrix and $T$ is a $3 \times 1$ translation matrix. For any point in the world coordinate system, the corresponding image pixel location can be found with the classic camera transformation \cite{hartley2003multiple}:
\begin{equation}
Y=K\times [R|T] \times X
\label{eq:camera-transform}
\end{equation}

where $X$ is a homogeneous world 3D coordinate of size $4 \times 1$, and $Y$ is a homogeneous 2D coordinate of size $3\times 1$.  Equation (\ref{eq:camera-transform}) is used for rendering models onto the image, as well as generating ground-truth labels that maps each vehicle's 3D bounding box in the image. In this work, the AR rendering task is accomplished with Pyrender, a light-weight AR rendering module for Python \cite{pyrender}. Figure \ref{fig:reality-enhancement} (top row) shows some rendering results.

\subsection{Reality Enhancement}
\label{subsection:reality-enhancement}
The AR method from section \ref{subsection:augmented-reality} creates vehicles in the foreground over the real background images. These foreground vehicles are rendered from 3D models, which are not realistic enough and may negatively impact the real-world performance of the trained detector. To address this issue, we have incorporated a GAN-based reality enhancement component that converts the AR generated foreground vehicles to a more realistic appearance. Specifically, we have utilized the Contrastive Unpaired Translation (CUT) technique to translate the AR-generated foreground to a realistic image style \cite{park2020contrastive}. The realistic image styles have been learned from BAAI-Vanjee dataset \cite{abs-2105-14370}, which comprises $2000$ roadside camera images. To remove the backgrounds of all images in the dataset, we have employed a salient object detector known as TRACER \cite{lee2022tracer}. The vehicles in the images have been cropped according to the annotation, enabling the CUT model to focus solely on translating the vehicle style rather than the background style. The AR-rendered vehicles have been translated individually and re-rendered to the same position. Figure \ref{fig:reality-enhancement} displays some sample images. 
The top row of four images are the AR rendered images, the bottom row are the corresponding images after reality enhancement, arranged in the order of
eastbound approach,
northbound approach,
southbound approach
and westbound approach camera views.

\subsection{Vehicle Detection and Localization}
\label{subsection:detection-and-localization}
\begin{figure}[hb]
    \vspace{-5mm}
    \centering
    \includegraphics[width=1\linewidth]{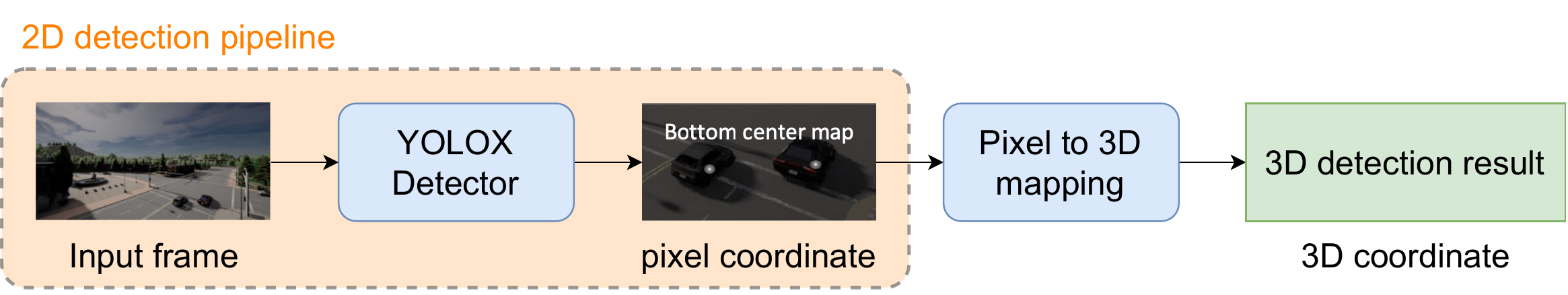}
    \caption{Complete detection pipeline for 3D vehicle localization and its relationship with our trained YOLOX Detector}
    \label{fig:detection-pipeline}
\end{figure}
In the applications for autonomous vehicles, vehicle 3D coordinates are needed. This requires an extra step that lifts the 2D detection results to 3D. This paper adopts a previously reported method to achieve this goal \cite{zhang2022design, zou2022real}. Figure \ref{fig:detection-pipeline} illustrates an overall pipeline for estimating vehicle 3D coordinates. The 2D detector is trained to detect the vehicles' bottom centers in the image. These bottom center points are mapped with a homography map to the road plane. The homography matrix can be obtained from the same correspondence set between the real-world and image described in section \ref{subsection:pose-estimation}. We use a YOLOX detector to perform the 2D bottom center detection task \cite{abs-2107-08430}. To train our YOLOX model to detect vehicle bottom centers, the data annotations generated in section \ref{subsection:augmented-reality} record bounding boxes that tightly frame the vehicles' 3D bottom in the image. In this way, one can recover the bottom center position from the YOLOX detection results simply by using the center point of the bounding box.  As this method has been thoroughly tested, reported and proven to be viable (\cite{zhang2022design, zou2022real}), and with accurate homography map, the 3D localization evaluation is equivalent to pixel localization performance. in this paper, our main focus is on 2D detection and pixel-level localization, framed with yellow box in figure \ref{fig:detection-pipeline}. While the detection and localization technologies are crucial components of our study, it is important to note that our primary focus is not on these technologies per se. The method we employ here, which is among the state-of-the-art in detection and localization, serves as a representative example. Our presented pipeline is designed to be universally compatible with a variety of other advanced methods, including but not limited to YOLO \cite{redmon2016you}, SSD \cite{liu2016ssd}, and Faster-RCNN \cite{ren2015faster}.

\subsection{Deployment Strategy}
 As previously discussed, the roadside vehicle detection strategy presented in this paper lessens the dependence on human labeling. The deployment strategy is as follows:

\begin{enumerate} [label=Step \arabic*:, leftmargin=*]
\item Install cameras at the roadside, determine the camera's intrinsic matrix, and perform pose estimation.
\item Following camera installation, begin collecting background images.
\item Superimpose rendered vehicles onto these collected background images, along with their corresponding labels, to create a synthesized dataset.
\item Use this synthesized dataset to train a deep learning-based detector. One might opt to initialize training using a pretrained detector from another location or dataset as a starting checkpoint.
\item Roll out the detector trained on the synthesized dataset.
\item Periodically amass additional background images to expand the synthesized dataset, thereby enhancing background diversity coverage. Iteratively refine the detector using this continually augmented synthesized dataset.
\end{enumerate}

It's noteworthy that, theoretically, this entire deployment pipeline could operate with full automation, allowing the detector to update and evolve autonomously without, or with minimum human intervention.

\begin{figure*}[t]
    \centering
    \includegraphics[width=.16\textwidth, trim={1mm, 0mm, 2mm, 0mm}, clip]{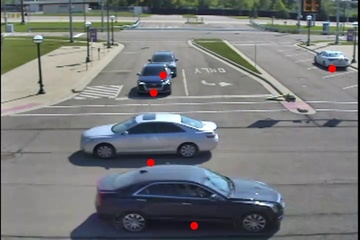}
    \includegraphics[width=.16\textwidth, trim={1mm, 0mm, 2mm, 0mm}, clip]{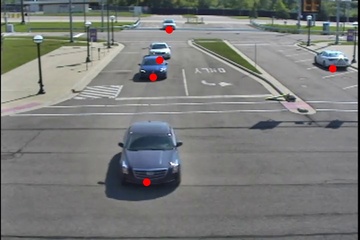}
    \includegraphics[width=.16\textwidth, trim={1mm, 0mm, 2mm, 0mm}, clip]{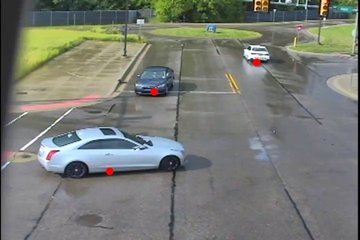}
    \includegraphics[width=.16\textwidth, trim={1mm, 0mm, 2mm, 0mm}, clip]{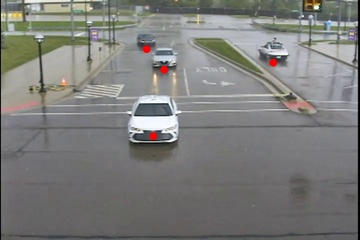}
    \includegraphics[width=.16\textwidth, trim={1mm, 0mm, 2mm, 0mm}, clip]{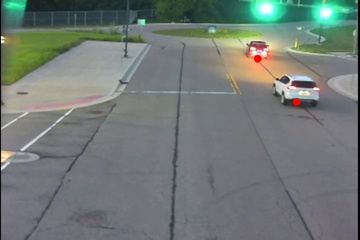}
    \includegraphics[width=.16\textwidth, trim={1mm, 0mm, 2mm, 0mm}, clip]{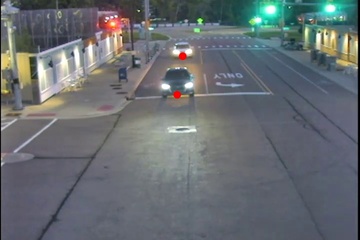}
    \caption{Visualization of detection results on various
    conditions (left two: sunny weather, center two: rainy weather,
    right two: dark lighting condition).
    The detector is trained on our synthesized dataset.
    {\color{red}Red} dots represent detected vehicle bottom centers.}
\vspace{-2mm}
    \label{fig:visualization-detection}
\end{figure*}

\begin{figure*}[t]
    \centering
    \includegraphics[width=.242\textwidth, trim={1mm, 0mm, 2mm, 0mm}, clip]{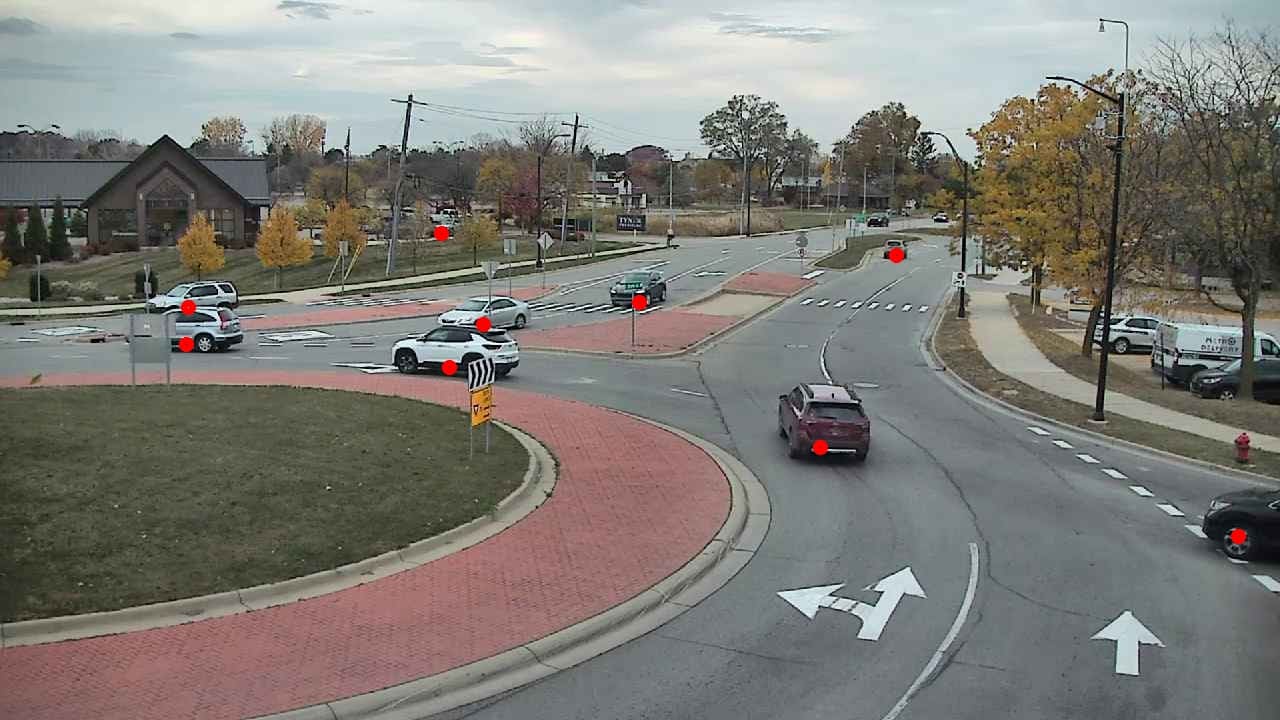}
    \includegraphics[width=.242\textwidth, trim={1mm, 0mm, 2mm, 0mm}, clip]{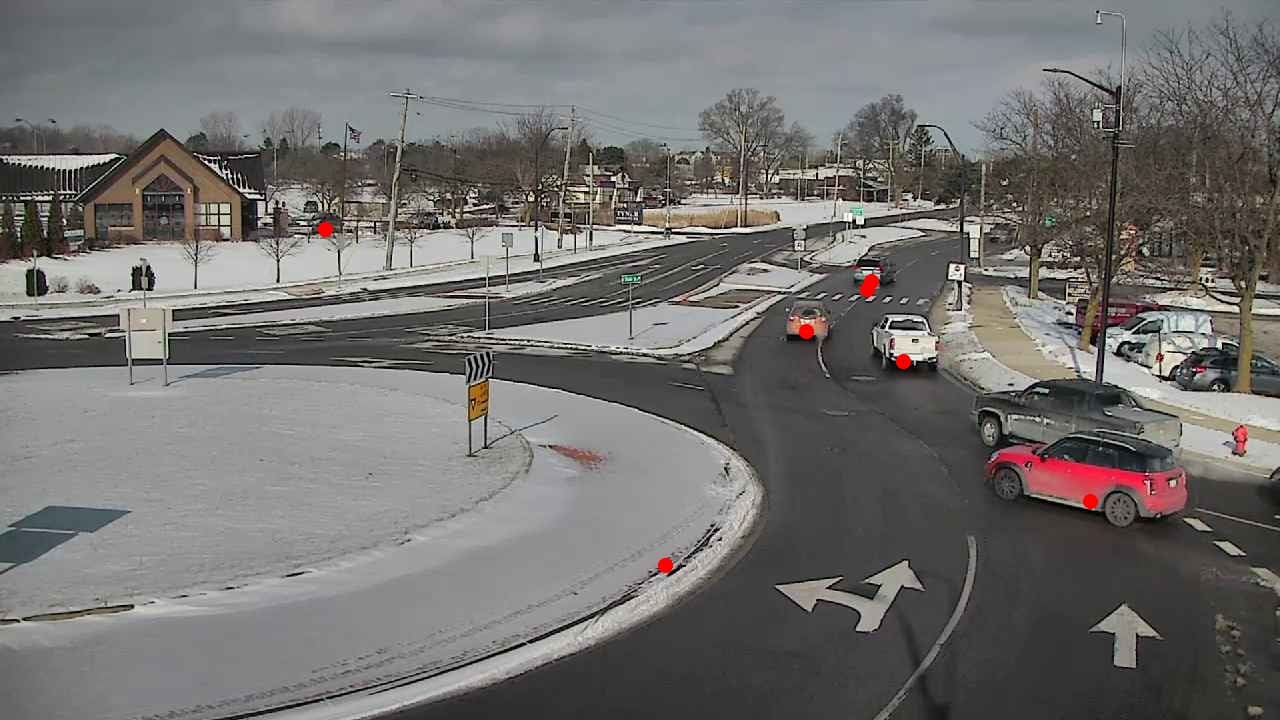}
    \includegraphics[width=.242\textwidth, trim={1mm, 0mm, 2mm, 0mm}, clip]{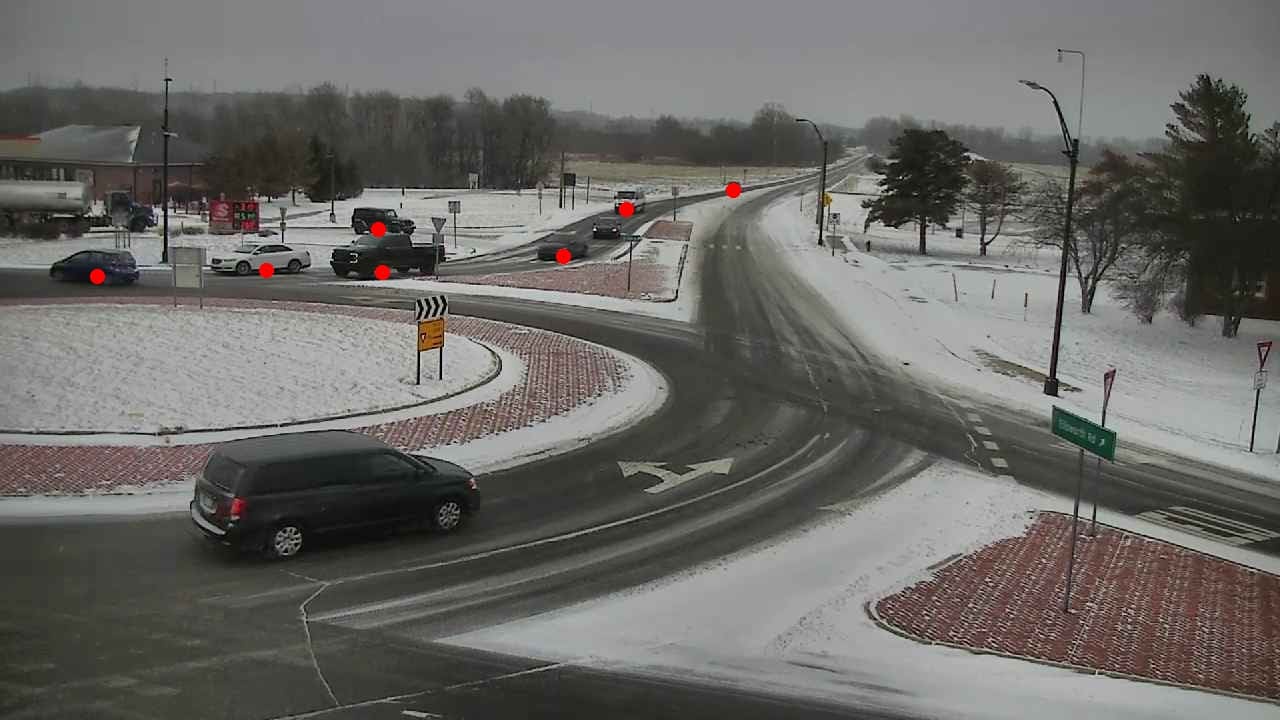}
    \includegraphics[width=.242\textwidth, trim={1mm, 0mm, 2mm, 0mm}, clip]{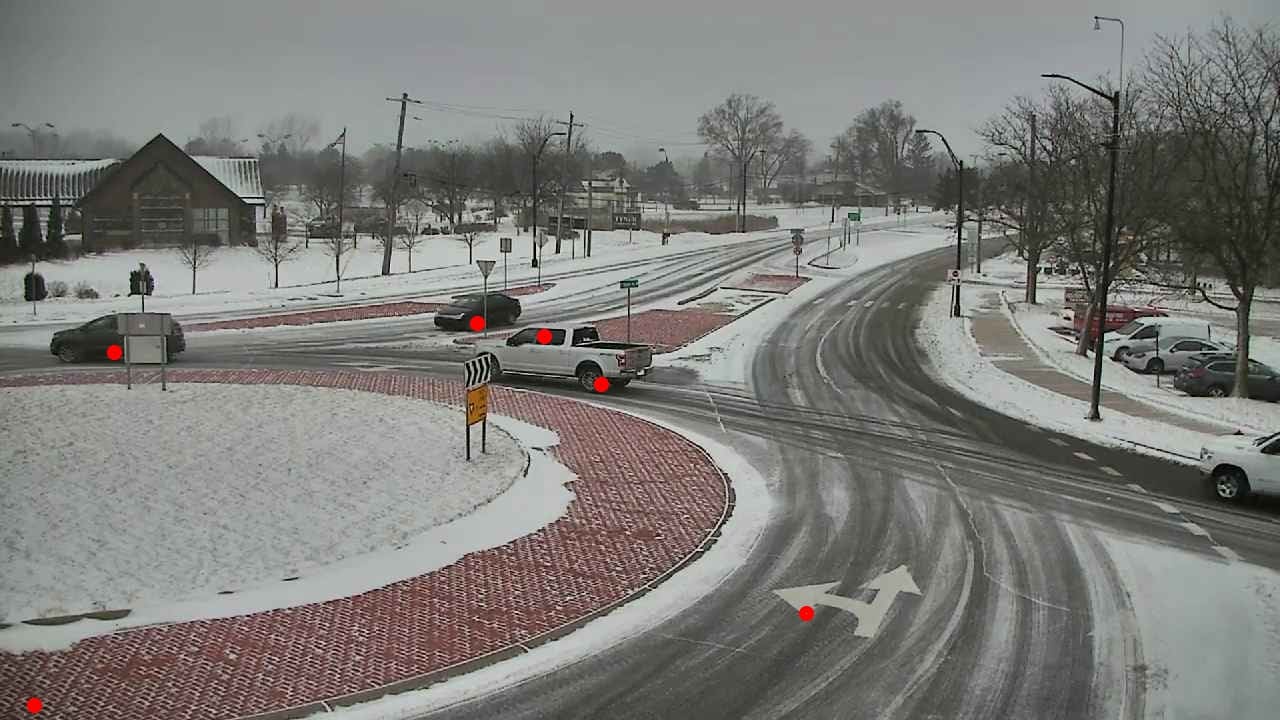} \\
    \vspace{1mm}\hspace{-1.00mm}
    \includegraphics[width=.242\textwidth, trim={1mm, 0mm, 2mm, 0mm}, clip]{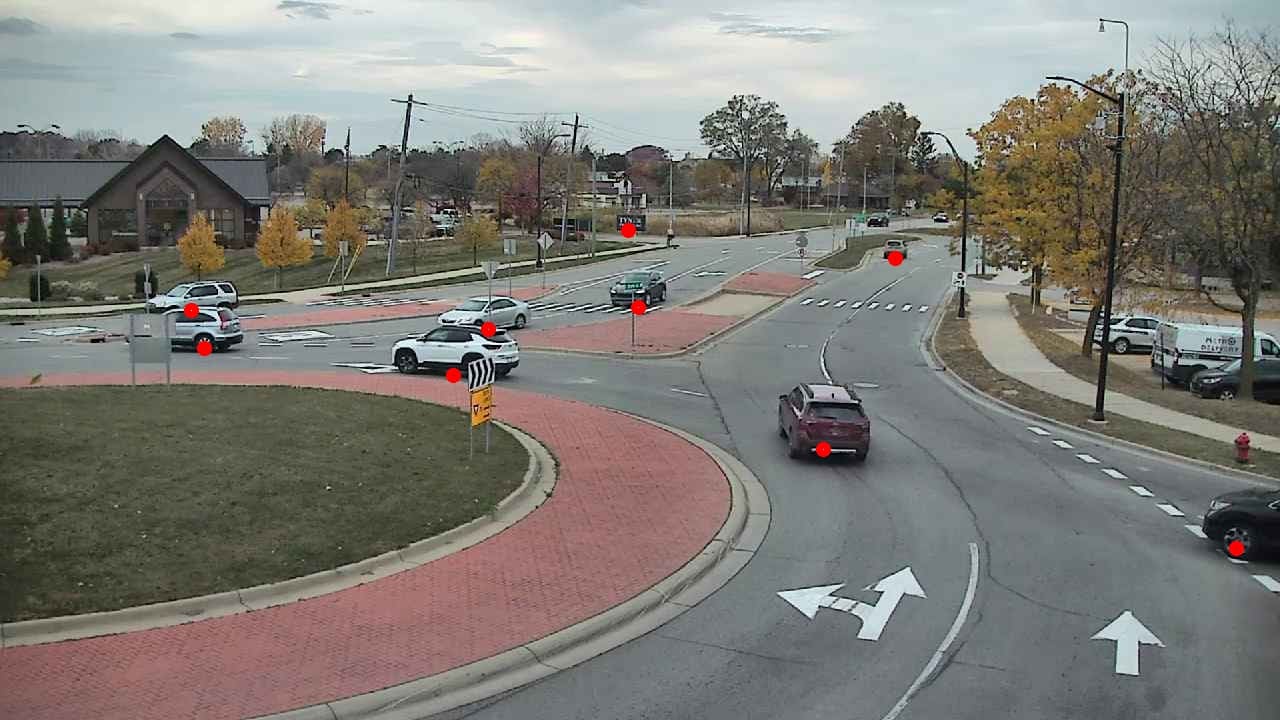}
    \includegraphics[width=.242\textwidth, trim={1mm, 0mm, 2mm, 0mm}, clip]{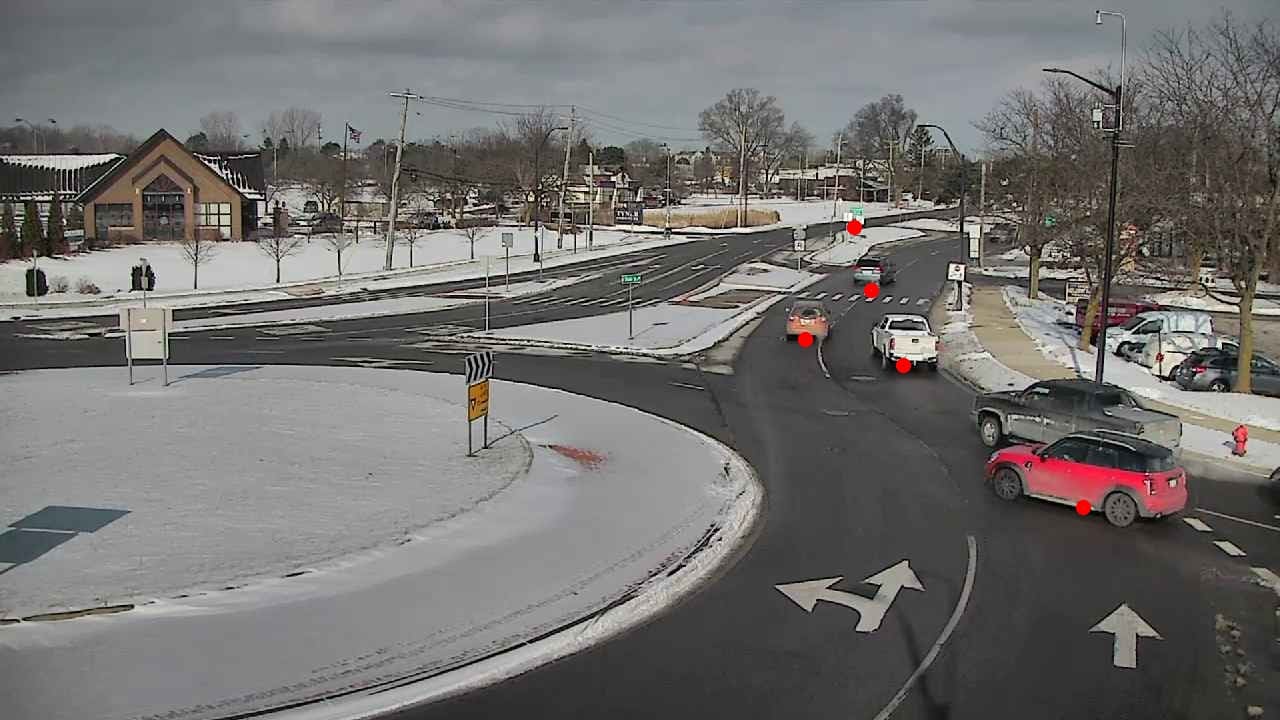}
    \includegraphics[width=.242\textwidth, trim={1mm, 0mm, 2mm, 0mm}, clip]{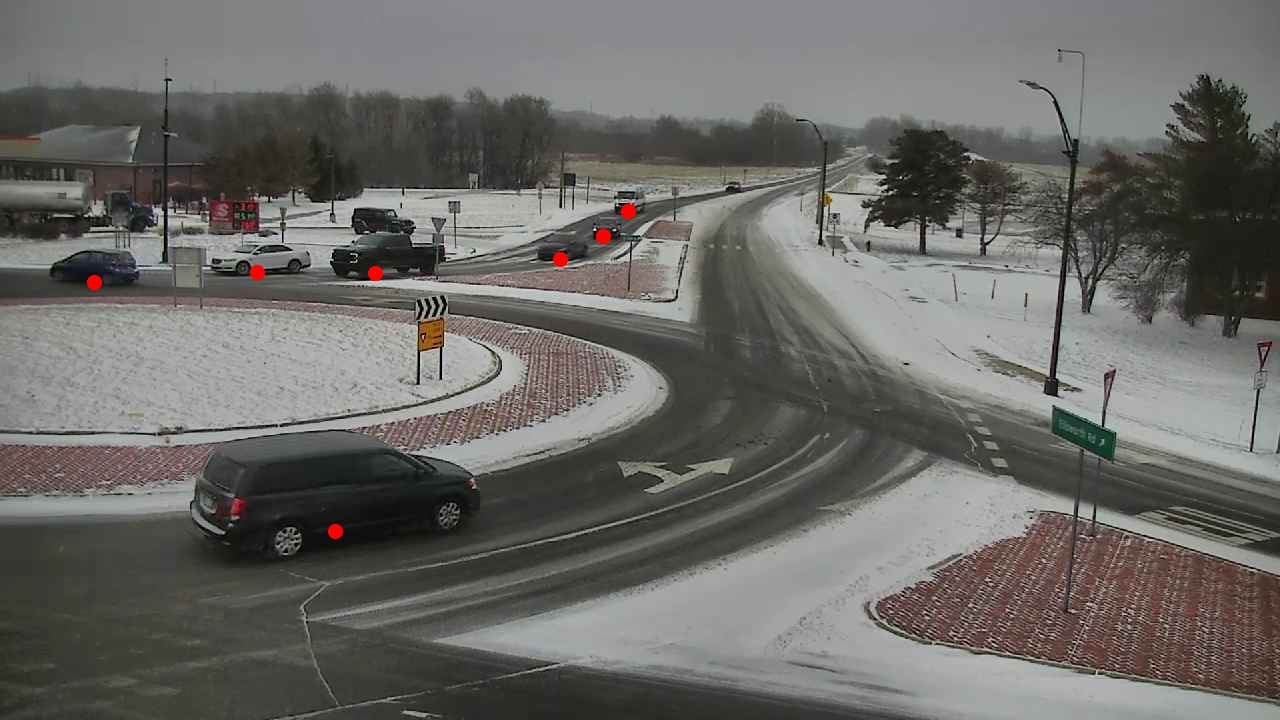}
    \includegraphics[width=.242\textwidth, trim={1mm, 0mm, 2mm, 0mm}, clip]{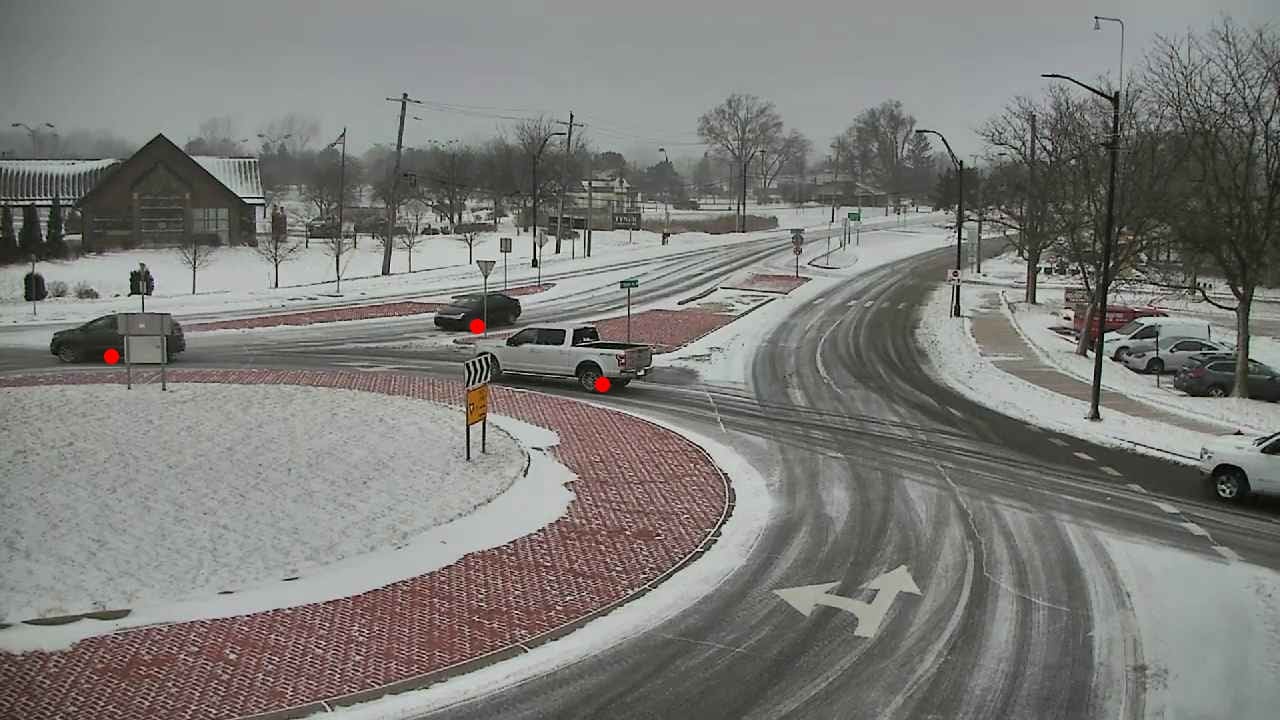}
    \caption{Comparison of detection results
    of the detector trained on human-labeled real data (top),
    and the detector trained on a mixed dataset including human-labeled data
    and synthesized data (bottom) at the State Street/Ellsworth Road intersection.
    Synthesized data improves the detection quality in harsh conditions (snow weather).
    {\color{red}Red} dots represent detected vehicle bottom centers.}
\vspace{-5mm}
    \label{fig:visualization-detection-rb}
\end{figure*}
\section{Performance Evaluation}
\label{section:performance-evaluation}
\subsection{Training Dataset}
\noindent\textbf{Mcity Intersection}
Mcity is a closed testing facility that does not have urban traffic flow.
Therefore, we use pure synthesized images for the detector training.
Our training dataset contains $4,000$ images in total.
We synthesize $1,000$ images for each camera view (north, south, east and west).
The background images for synthesis are captured
and sampled from roadside camera clips
with $720\times 480$ resolution
over $5$ days. 
For the foreground, we consider all kinds of
vehicles (cars, buses, trucks, etc)
to be in the same 'vehicle' category. The backgrounds for synthesizing data and those used in the final evaluation are intentionally selected from different days to ensure no repetition. We choose similar weather conditions and made the backgrounds resemble those in the evaluation dataset. We introduce uniformly sample $100$ background images
from $8$am to $8$pm for data synthesizing.

\noindent\textbf{State St/Ellsworth Rd Roundabout}
We also tested our approach on the roundabout at 
State Street and Ellsworth Road at Ann Arbor, Michigan.
A perception system trained with $1$K manually labeled images has already been deployed in fall (September, 2022). The performance of which decades in winter (December, 2022).
Therefore, we use our synthesis method to generate $1$K more
images to improve the performance in the winter.
The background images for synthesis are captured 
with different conditions (sunny, cloudy, snow) from different days.

\subsection{Evaluation Metrics}
Our evaluation method aligns closely with the standard approach used in general object detection \cite{everingham2010pascal}. However, traditional evaluation methods do not sufficiently emphasize localization precision, a critical factor in cooperative driving. Drawing inspiration from the evaluation metrics proposed by nuScenes~\cite{CaesarBLVLXKPBB20} and CLEAR metrics \cite{bernardin2008evaluating}, we have developed an evaluation method similar to general object detection, with a key distinction: we identify false positives using a distance threshold.

Specifically, our metrics are based on the pixel distance between the bottom centers of vehicles. We calculate the center distance between the detected vehicle and ground truth, denoted as \(d\). We set a distance error tolerance of \(\theta\) and consider detections with \(d < \theta\) as true positives and those with \(d \geq \theta\) as false positives. Detections are sorted in descending order of confidence scores for Average Precision (AP) calculation. We calculate AP for \(\theta=2, 5, 10, 15, 20, 50\) pixels, as well as the mean average precision (mAP). We report mAP, AP@\(20\) (AP with \(\theta=20\) pixels), AP@\(50\) (AP with \(\theta=50\) pixels), and the average recall (AR).

\subsection{Training Settings}
\label{subsection:training-settings}
We follow the training pipeline provided by YOLOX~\cite{abs-2107-08430}
with some modifications to fit our dataset.
We use YOLOX-Nano as the default model in our
experiments. We train the model for $150$ epochs
in total with $15$ warm-up epochs included,
and drop the learning rate by a factor of $10$ after $100$ epochs.
The initial learning rate is set to be $4e-5$
and the weight decay is set to be $5e-4$.
The Adam~\cite{KingmaB14} optimizer is used.
We train the model with a mini-batch size $8$ on one NVIDIA RTX $3090$ GPU.

For data augmentation, we first resize the input
image such that the long side is at $640$ pixels,
and then pad the short side to $640$ pixels.
Random horizontal flips are applied with probability $0.5$
and a random Hue, Saturation, and Value (HSV) augmentation is applied with a
gain range of $[5, 30, 30]$.

\subsection{Experiments and Evaluation at Mcity}
To thoroughly test the robustness of the proposed perception system, a total of six field-tests were carried out at Mcity during the months of July and August, 2022. During the field tests, vehicles adhered to traffic and lane rules while traversing the intersection for at least 15 minutes per trial. To ensure an adequate level of diversity, over 20 distinct vehicles were used in the experiment. These six trials encompassed a wide range of environmental variations including different weather (sunny, cloudy, light raining, heavy raining) and lighting (daytime and nighttime) conditions. 

Two evaluation datasets were built from the field tests described above:
normal condition evaluation dataset
and harsh condition evaluation dataset.
The normal condition dataset contains $217$ images
with real vehicles in the intersection during
the daytime under good weather conditions.
The harsh condition dataset contains $134$ images
with real vehicles in the intersection
under adverse conditions.
$15$ images are under light raining conditions,
$39$ images are collected at twilight or dusk,
$50$ images are collected under heavy raining conditions,
$30$ images are collected in sunshine after raining conditions.

\begin{table*}[t]
\centering
\renewcommand{\arraystretch}{1.3}
\setlength{\tabcolsep}{11pt}
 \begin{tabular}{l|c|cccc|cccc} 
 \shline
 \multirow{2}{*}{Training dataset} & \multirow{2}{*}{\#images} & \multicolumn{4}{c|}{Normal Condition Evaluation} & \multicolumn{4}{c}{Harsh Condition Evaluation} \\ 
 \cline{3-10}
 & & mAP & AP@$20$ & AP@$50$ & AR 
 & mAP & AP@$20$ & AP@$50$ & AR \\
 \shline
 COCO~\cite{LinMBHPRDZ14} & $118$K & $47.5$ & $70.3$ & $88.9$ & $62.1$ & $38.3$ & $54.4$ & $85.2$ & $57.6$ \\  
 KITTI~\cite{Geiger2012CVPR} & $8$K & $46.4$ & $76.2$ & $89.5$ & $62.8$ & $33.6$ & $54.7$ & $75.8$ & $53.6$ \\
 BAAI-Vanjee~\cite{abs-2105-14370} & $2$K & $42.5$ & $65.3$ & $84.9$ & $62.6$ & $34.7$ & $48.7$ & $80.6$ & $57.2$ \\  
 DAIR-V2X~\cite{abs-2204-05575}\hspace{10mm} & $7$K & $39.7$ & $60.1$ & $71.6$ & $60.1$ & $34.1$ & $51.0$ & $62.4$ & $54.3$ \\
 \hline
 Ours & $4$K & $\mathbf{49.1}$ & $\mathbf{78.0}$ & $\mathbf{92.4}$ & $\mathbf{63.6}$ & $\mathbf{44.4}$ & $\mathbf{72.1}$ & $\mathbf{89.8}$ & $\mathbf{59.1}$ \\
 \shline
 \end{tabular}
\vspace{-2mm}
 \caption{Comparison of model trained on our synthesis dataset
 to models on other existing datasets.
 The model trained on our synthesis dataset achieves the
 best performance on both normal conditions and harse conditions.}
\label{table:baseline}
\end{table*}

\begin{table*}[t]
\centering
\renewcommand{\arraystretch}{1.3}
\setlength{\tabcolsep}{9.3pt}
 \begin{tabular}{cccc|cccc|cccc} 
 \shline
 \multicolumn{4}{c|}{Settings} & \multicolumn{4}{c|}{Normal Condition Evaluation} & \multicolumn{4}{c}{Harsh Condition Evaluation} \\ 
 \hline
 AR & AR + RE & Single bg. & Diverse bg. & mAP & AP@20 & AP@50 & AR & mAP 
 & AP@20 & AP@50 & AR \\
 \shline
 \cmark & & \cmark & & $34.8$ & $63.0$ & $84.8$ & $54.4$ & $29.9$ & $50.1$ & $77.7$ & $49.8$ \\  
 \cmark & & & \cmark & $40.1$ & $66.1$ & $88.5$ & $57.9$ & $37.0$ & $62.7$ & $85.8$ & $53.9$  \\ 
 & \cmark & \cmark & & $43.4$ & $73.4$ & $89.1$ & $57.4$ & $37.1$ & $64.4$ & $82.4$ & $54.8$ \\
 & \cmark & & \cmark & $49.1$ & $78.0$ & $92.4$ & $63.6$ & $44.4$ & $72.1$ & $89.8$ & $59.1$ \\
 \shline
 \end{tabular}
 \caption{Ablation study. In the settings, \textbf{AR} means to directly use
 Augmented Reality to render vehicles. \textbf{AR+RE} means
 to use Augmented Reality with Reality Enhancement for vehicle
 generation. \textbf{Single bg.} means to use only one single
 background for dataset generation.
 \textbf{Diverse bg.} means to use diverse backgrounds
 for dataset generations.}
\label{table:ablation-components}
\end{table*}

We compare YOLOX-Nano trained on our synthesized data
to the same model trained on other datasets, including
the general object detection dataset COCO~\cite{LinMBHPRDZ14},
the vehicle-side perception dataset KITTI~\cite{Geiger2012CVPR},
and the roadside perception datasets BAAI-Vanjee~\cite{abs-2105-14370}
and DAIR-V2X~\cite{abs-2204-05575}. 
Since we evaluate the vehicle bottom center position,
while the following datasets only provide the object bounding box
in their $2$d annotations,
for models trained on COCO, KITTI, BAAI-Vanjee, and DAIR-V2X,
we manually apply a center shift to roughly map
the predicted vehicle center to vehicle bottom center by
$x_{bottom} = x, y_{bottom} = y + 0.35h$.
Here $(x_{bottom}, y_{bottom})$ is the estimated vehicle bottom
center after mapping,
and the $(x, y)$ is the predicted object center by the detector.

Table~\ref{table:baseline} shows the comparison between the model
trained on our dataset and on other datasets.
We use the pretrained YOLOX weight as the initial weights of our training process, this is a standard practice for YOLOX detector fine-tuning.
The model trained on our dataset outperforms
all other datasets on both normal conditions
and harsh conditions.
On normal conditions, our model achieves $1.6$ mAP improvement
and $1.5$ AR improvement over the second best model (trained on COCO).
On harsh conditions, our model achieves $6.1$ mAP improvement
and $1.5$ AR improvement over the model trained on COCO. One can observe that our method enhances performance under both normal and harsh conditions, with a more significant improvement noted in harsh conditions. This improvement is due to the lack of harsh condition data in the original training dataset, which our synthesized dataset compensates for. Notably, the performance enhancement is greater at a 20-pixel threshold than at a 50-pixel threshold. This indicates that our method effectively guides the detector to improve localization accuracy, a critical aspect in the context of cooperative driving.

For other datasets,
one can see that the models trained on roadside perception
datasets (BAAI-Vanjee and DAIR-V2X) are worse than 
COCO and KITTI on normal conditions.
This implies that the roadside perception datasets
might have a weaker transfer-ability than
general object detection datasets.
One possible reason might be the poses of the camera
are fixed.
On harsh conditions, none of the existing datasets achieve
satisfactory performance.

\begin{table}[t]
\centering
\renewcommand{\arraystretch}{1.3}
\setlength{\tabcolsep}{9pt}
 \begin{tabular}{l|c|c|c} 
 \shline
 \multirow{1}{*}{Training dataset} & \multirow{1}{*}{\#images} & Normal & Snow \\
 \shline
 Real data (human labeled) & $1$K & $51.3$ & $39.7$ \\  
 Mixed (real + synthesized) & $1$K + $1$K & $53.6$ & $45.8$ \\  
 \shline
 \end{tabular}
 \caption{Comparison of model trained on 
 $1$K labeled data and model trained on the
 dataset of mixed labeled and synthesized data.
 mAP is reported.}
\label{table:rb}
\end{table}

\begin{table}[t]
\centering
\renewcommand{\arraystretch}{1.3}
\setlength{\tabcolsep}{5pt}
 \begin{tabular}{cc|cc|cc} 
 \shline
 \multicolumn{2}{c|}{Diversity of backgrounds} & \multicolumn{2}{c|}{Normal} & \multicolumn{2}{c}{Harsh} \\ 
 \hline
 Weather diversity & Time diversity & mAP & AR 
 & mAP & AR \\
 \shline
 \xmark & \xmark & $43.4$ & $57.4$ & $37.1$ & $54.8$ \\
 \cmark & \xmark & $46.8$ & $54.7$ & $40.1$ & $57.3$ \\
 \xmark & $1$ day, $8$am to $8$pm & $47.4$ & $60.3$ & $41.8$ & $56.8$ \\
 \cmark & $5$ day, $8$am to $8$pm & $49.1$ & $63.6$ & $44.4$ & $59.1$ \\
 \shline
 \end{tabular}
 \caption{Ablation study on diversity of backgrounds. 
 Adding weather diversity and time diversity both improve
 the detection performance on all conditions.
 Improvement on harsh conditions is more significant.}
\label{table:ablation-background}
\vspace{-6mm}
\end{table}
\subsection{Experiments at State St/Ellsworth Rd Roundabout}
As mentioned earlier, a roadside perception system has already been deployed~\cite{zou2022real} with training data
collected and annotated in the fall.
It has been found that the perception system degrades dramatically during winter, especially in snowy conditions.
To mitigate this issue, we use the synthesizing pipeline introduced in this paper to enrich training samples with winter and fall background.

For evaluation, we collected $343$ image data from the roundabout site in normal weather conditions and snow weather conditions.
For normal weather conditions, we collected and annotated $111$ images.
For snow weather conditions, $232$ images are contained for evaluation.

As shown in Table~\ref{table:rb},
We compare the model trained with real data only
(collected in the fall),
and the model trained with mixed data (real data collected
in fall, and synthesized data collected in both fall and
winter). The model trained on mixed data
outperforms the model trained on real data on both
normal condition and snowy weather condition.
Under snowy conditions, the improvement of $6.1$ mAP
is obtained.
Figure~\ref{fig:visualization-detection-rb}
shows the comparison of the two models.
Both detectors perform decently in the fall. Under snowy weather conditions,
the model trained with mixed data performs better.

\subsection{Ablation Study}

\vspace{1mm}
\noindent\textbf{Analysis on components of the pipeline.}
In our data synthesis analysis, we focus on two key components: GAN-based reality enhancement (RE) and diverse backgrounds. In Table~\ref{table:ablation-components}, we compare four settings: AR only with a single background, AR only with diverse backgrounds, AR + RE with a single background, and AR + RE with diverse backgrounds. The 'diverse background' refers to the shuffled selection from multiple backgrounds used in our training dataset, as mentioned in previous section. In contrast, the 'single background' dataset consistently uses the same background for synthesizing data. We observe that applying diverse backgrounds to AR-only data improves mAP by $5.3$ in normal conditions and $7.1$ in harsh conditions. Additionally, when compared to AR-only with a single background, adding RE leads to an improvement of $8.6$ mAP in normal conditions and $7.3$ mAP in harsh conditions. The combination of diverse backgrounds and reality enhancement further enhances performance by over $5$ mAP in normal conditions and $7$ mAP in harsh conditions.
\vspace{1mm}

\noindent\textbf{Analysis on diversity of backgrounds.}
Using diverse backgrounds in image rendering is the key
to achieve robust vehicle detection over different
lighting conditions and weather conditions.
Table~\ref{table:ablation-background}
shows the analysis on diversity of backgrounds.
We can clearly observe that both weather diversity and time diversity improve
the detection performance.
An interesting finding is that the performance
on normal conditions
is also greatly improved by the diverse backgrounds.

\begin{table}[t]
\centering
\renewcommand{\arraystretch}{1.3}
\setlength{\tabcolsep}{9pt}
 \begin{tabular}{l|cc|cc} 
 \shline
 \multirow{2}{*}{Pretrain dataset} & \multicolumn{2}{c|}{Normal} & \multicolumn{2}{c}{Harsh} \\ 
 \cline{2-5}
 & mAP & AR 
 & mAP & AR \\
 \shline
 $-$ & $43.7$ & $63.8$ & $34.7$ & $60.6$ \\
  KITTI~\cite{Geiger2012CVPR}  & $48.2$ & $66.2$ & $41.6$ & $60.4$ \\
  BAAI-Vanjee~\cite{abs-2105-14370}\hspace{10mm} & $42.4$ & $59.0$ & $40.6$ & $56.3$ \\
  DAIR-V2X~\cite{abs-2204-05575} & $44.8$ & $61.7$ & $40.3$ & $58.1$ \\
  COCO~\cite{LinMBHPRDZ14}  & $49.1$ & $63.6$ & $44.4$ & $59.1$ \\
 \shline
 \end{tabular}
 \caption{Ablation study on pretraining. Pretraining on existing datasets
 improves mAP on both normal conditions and harsh conditions. AR is not 
 improved by pretraining.}
\label{table:ablation-pretrain}
\end{table}

\vspace{1mm}
\noindent\textbf{Analysis on the Impact of Different Pretrained Weights.}
In Table~\ref{table:ablation-pretrain},
we show that our method can also benefit from
pretraining on existing datasets.
On normal conditions,
pretraining on COCO dataset or KITTI dataset
improves the detection performance by over $4$ mAP,
while pretraining on BAAI-Vanjee or DAIR-V2X
dataset shows no significant improvement.
One possible reason is that the BAAI-Vanjee
dataset and DAIR-V2X dataset
are roadside datasets captured in Chinese intersections.
The generalization ability to U.S. intersections
might be limited.
On harsh conditions, pretraining on all datasets
shows decent mAP improvement.

\section{Discussion and Future work}
\label{section:discussion}
From section \ref{section:performance-evaluation}, it can be observed that the model's performance is enhanced after tuning on our synthesized dataset, especially in terms of precision under harsh conditions. It is worth noting that the improvement in recall is relatively marginal in most cases. An intuitive explanation can be provided here. By incorporating a substantial number of background images into the training dataset, the model is able to rectify instances where it incorrectly identifies backgrounds as vehicles. However, in order to enhance recall, the model must address cases where it misclassifies vehicles as backgrounds. In our specific case, there still exists a disparity between the synthesized vehicles in our dataset and real-world vehicles. As part of future work, we intend to enhance the quality of realistic vehicle synthesis. One promising direction is the utilization of Stable Diffusion \cite{rombach2022high} to render realistic images and achieve style translation. To be efficient, this approach will require a substantially larger dataset. Such a dataset could be feasibly acquired post-deployment in the real world, as part of the Smart Intersection Project (SIP) \cite{sip}. Futhremore, We plan to extend the investigation on more practical factors such as traffic flow conditions, shadows, and lighting, enriching the understanding on how different factors influence the performance.

\section{Conclusion}
In this study, we introduce a groundbreaking AR and GAN-based data synthesis pipeline, specifically designed to tackle the prevalent and crucial challenge of data insufficiency in current roadside vehicle perception systems. This innovative pipeline facilitates the training or fine-tuning of detectors, enabling them to seamlessly adapt to new locations, weather conditions, or lighting scenarios with minimal or virtually no human manual effort. This approach marks a significant leap forward in the real-world deployment of roadside detectors for autonomous driving, especially in the context of large-scale, robust deployment. A comprehensive evaluation is performed at different locations, under multiple weather and lighting conditions is reported in this paper. We demonstrate that our synthesized dataset can train a detector from scratch or fine-tune detectors trained from other datasets and improve the precision and recall under multiple lighting and weather conditions, yielding a much more robust perception system.

\bibliographystyle{IEEEtran}
\bibliography{IEEEabrv,bibfile}

\section{Biography}
\begin{IEEEbiography}[{\includegraphics[width=1in,height=1.25in,clip,keepaspectratio]{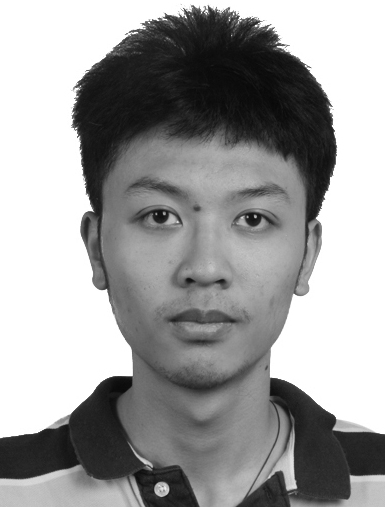}}]{Rusheng Zhang}
 received the B.E. degree in micro electrical mechanical system and second B.E. degree in Applied Mathematics from Tsinghua University, Beijing, in 2013. He received an M.S. and phD degree in electrical and computer engineering from Carnegie Mellon University, in 2015, 2019 respectively. His research areas include artificial intelligence, cooperative driving, cloud computing and vehicular networks.
\end{IEEEbiography}
\begin{IEEEbiography}[{\includegraphics[width=1in,height=1.25in,clip,keepaspectratio]{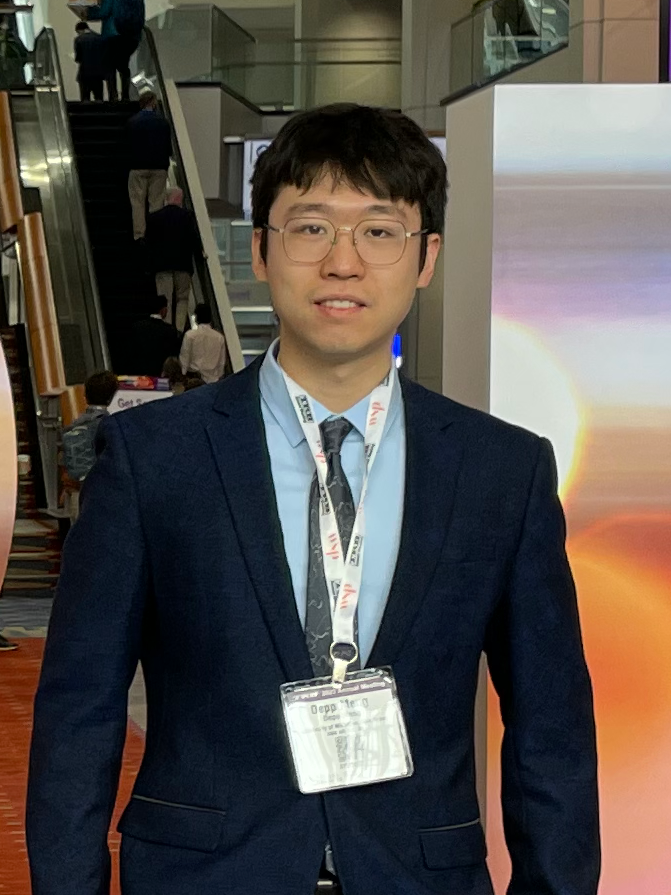}}]{Depu Meng} (Member, IEEE) is a Post Doctoral Research Fellow at the Department of Civil and Environmental Engineering, University of Michigan. He received his B. E. degree from the Department of Electrical Engineering and Information Science at the University of Science and Technology of China in 2018. He received his Ph. D. degree from the Department of Automation at the University of Science and Technology of China. His research interests include computer vision and autonomous driving systems.
\end{IEEEbiography}
\begin{IEEEbiography}[{\includegraphics[width=1in,height=1.25in,clip,keepaspectratio]{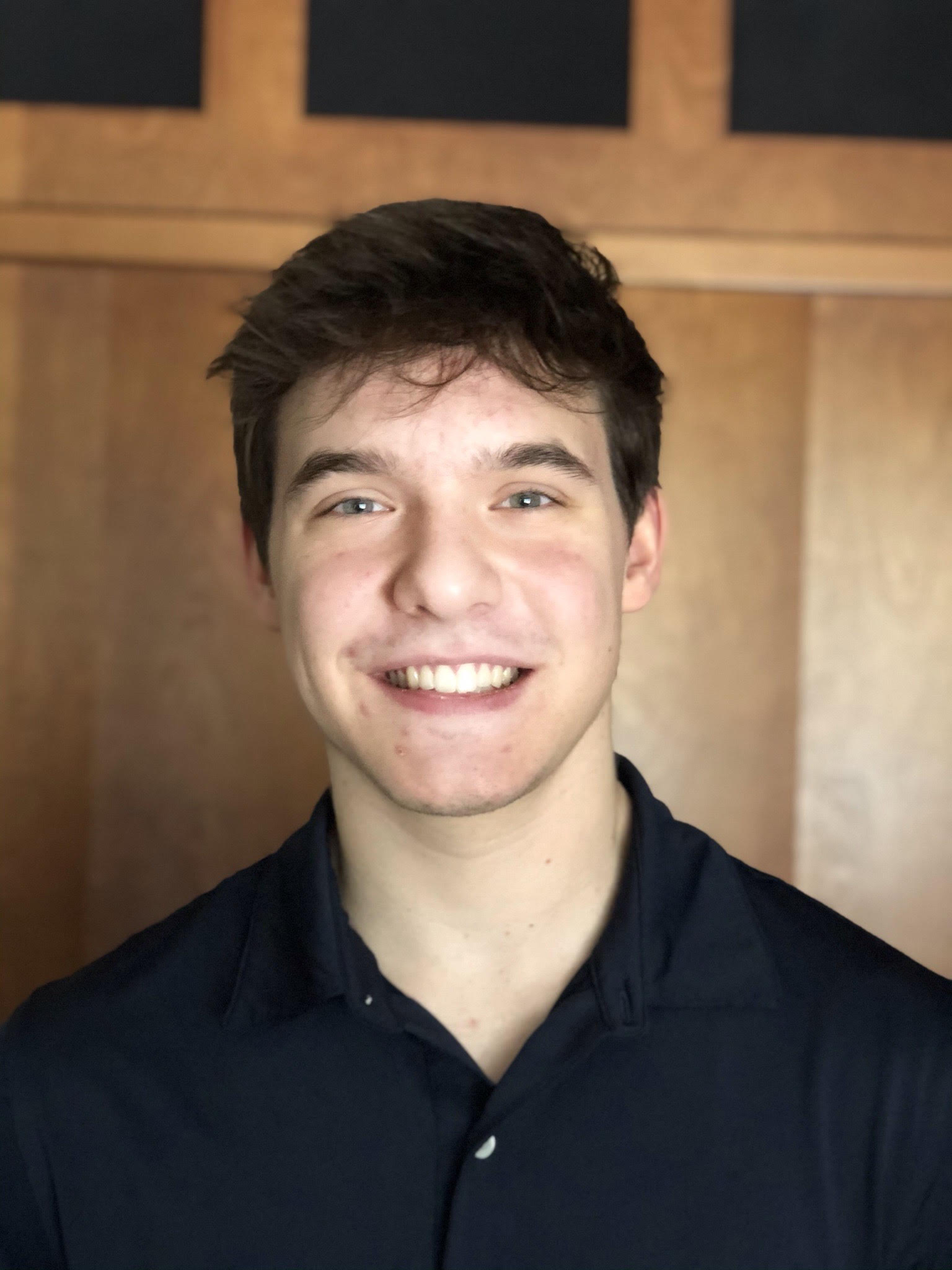}}]{Lance Bassett}  received a B.S. and M.S.E in Computer Science from the University of Michigan, Ann Arbor in 2023. He did research in computer vision with the Michigan Traffic Lab as a graduate student, and currently works on real-time software and media in the automotive
industry.
\end{IEEEbiography}
\begin{IEEEbiography}[{\includegraphics[width=1in,height=1.25in,clip,keepaspectratio]{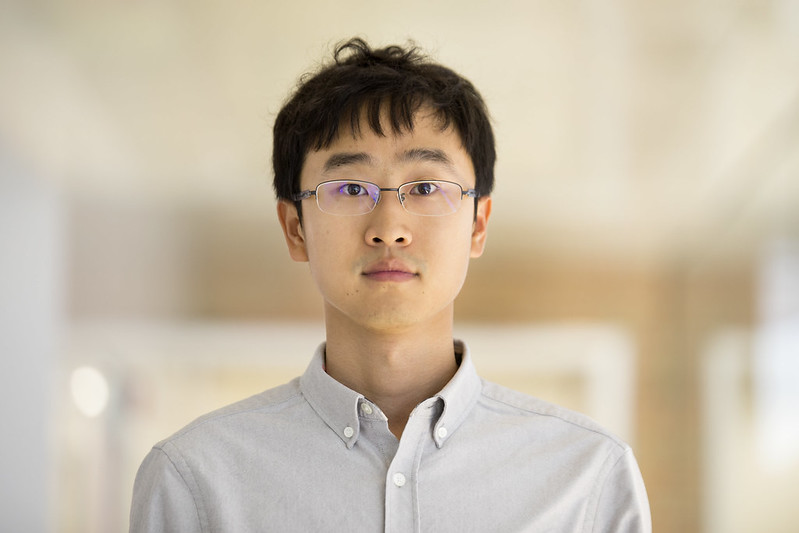}}]{Shengyin (Sean) Shen} works as a Research Engineer in the Engineering Systems Group at the University of Michigan Transportation Research Institute (UMTRI). 
Sean holds an MS degree in Civil and Environmental Engineering from the University of Michigan, Ann Arbor, and an MS degree in Electrical Engineering from the University of Bristol, UK. He also earned a BS degree from Beijing University of Posts and Telecommunications, China. Sean's research interests are primarily focused on cooperative driving automation and related applications that use roadside perception, edge-cloud computing, and V2X communications to accelerate the deployment of automated vehicles. 
He has extensive experience in implementation of large-scale deployments, such as the Safety Pilot Model Deployment (SPMD), Ann Arbor Connected Vehicle Testing Environment (AACVTE), and Smart Intersection Project. Moreover, he has been involved in many research projects funded by public agencies such as USDOT, USDOE, and companies such as Crash Avoidance Metric Partnership (CAMP), Ford Motor Company, and GM Company, among others.
\end{IEEEbiography}
\begin{IEEEbiography}
[{\includegraphics[width=1in,height=1.25in,clip,keepaspectratio]{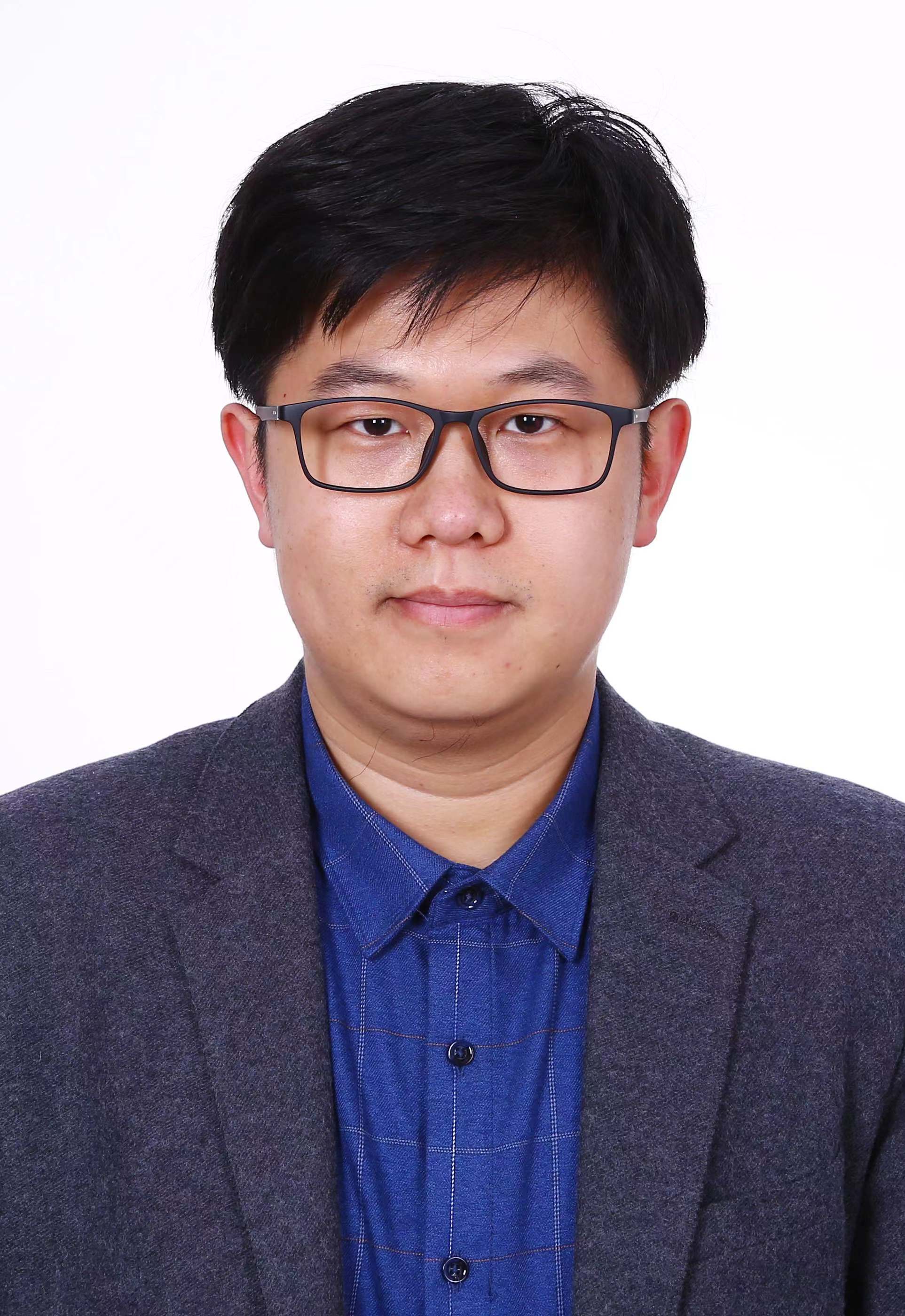}}]{Zhengxia Zou}
Zhengxia Zou received his B.S. and Ph.D. degrees from the Image Processing Center, School of Astronautics, Beihang University, Beijing, China, in 2013 and 2018, respectively. He is currently a Professor at the School of Astronautics, Beihang University. From 2018 to 2021, he worked at the University of Michigan, Ann Arbor as a Post-Doctoral Research Fellow. His research interests include computer vision and related problems in autonomous driving and remote sensing. He has published more than 40 peer-reviewed papers in top-tier journals and conferences, including Nature, Nature Communications, PROCEEDINGS OF THE IEEE, IEEE TRANSACTIONS ON IMAGE PROCESSING, IEEE TRANSACTIONS ON GEOSCIENCE AND REMOTE SENSING, and IEEE/CVF Computer Vision and Pattern Recognition. Dr. Zou was selected as “World’s Top 2\% Scientists” by Stanford University in 2022.
\end{IEEEbiography}
\begin{IEEEbiography}[{\includegraphics[width=1in,height=1.25in,clip,keepaspectratio]{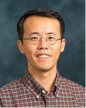}}]{Henry X. Liu}
(Member, IEEE) received the bachelor's degree in automotive engineering from Tsinghua University, China, in 1993, and the PhD. degree in civil and environment engineering from the University of Wisconsin-Madison in 2000. He is currently a professor in the Department of Civil and Environmental Engineering and the Director of Mcity at the University of Michigan, Ann Arbor. He is also a Research Professor at the University of Michigan Transportation Research Institute and the Director for the Center for Connected and Automated Transportation (USDOT Region 5 University Transportation Center). From August 2017 to August 2019, Prof. Liu served as DiDi Fellow and Chief Scientist on Smart Transportation for DiDi Global, Inc., one of the leading mobility service providers in the world. Prof. Liu conducts interdisciplinary research at the interface of transportation engineering, automotive engineering, and artificial intelligence. Specifically, his scholarly interests concern traffic flow monitoring, modeling, and control, as well as testing and evaluation of connected and automated vehicles. Prof. Liu is the managing editor of Journal of Intelligent Transportation Systems.
\end{IEEEbiography}

\vfill
\end{document}